%% file: main.tex
\documentclass[10pt,twocolumn,letterpaper]{article}

\usepackage{cvpr}              

\input{preamble}

\definecolor{cvprblue}{rgb}{0.21,0.49,0.74}
\usepackage[pagebackref,breaklinks,colorlinks,allcolors=cvprblue]{hyperref}

\def\paperID{44254} 
\def\confName{CVPR}
\def\confYear{2026}

\title{MusicInfuser: Making Video Diffusion Listen and Dance}

\author{Susung Hong
\and
Ira Kemelmacher-Shlizerman
\and
Brian Curless
\and
Steven M. Seitz
\and\\
University of Washington
}

\begin{document}

\twocolumn[{
\renewcommand\twocolumn[1][]{#1}
\maketitle
\begin{center}
\centering
\captionsetup{type=figure}
\includegraphics[width=0.9\textwidth]{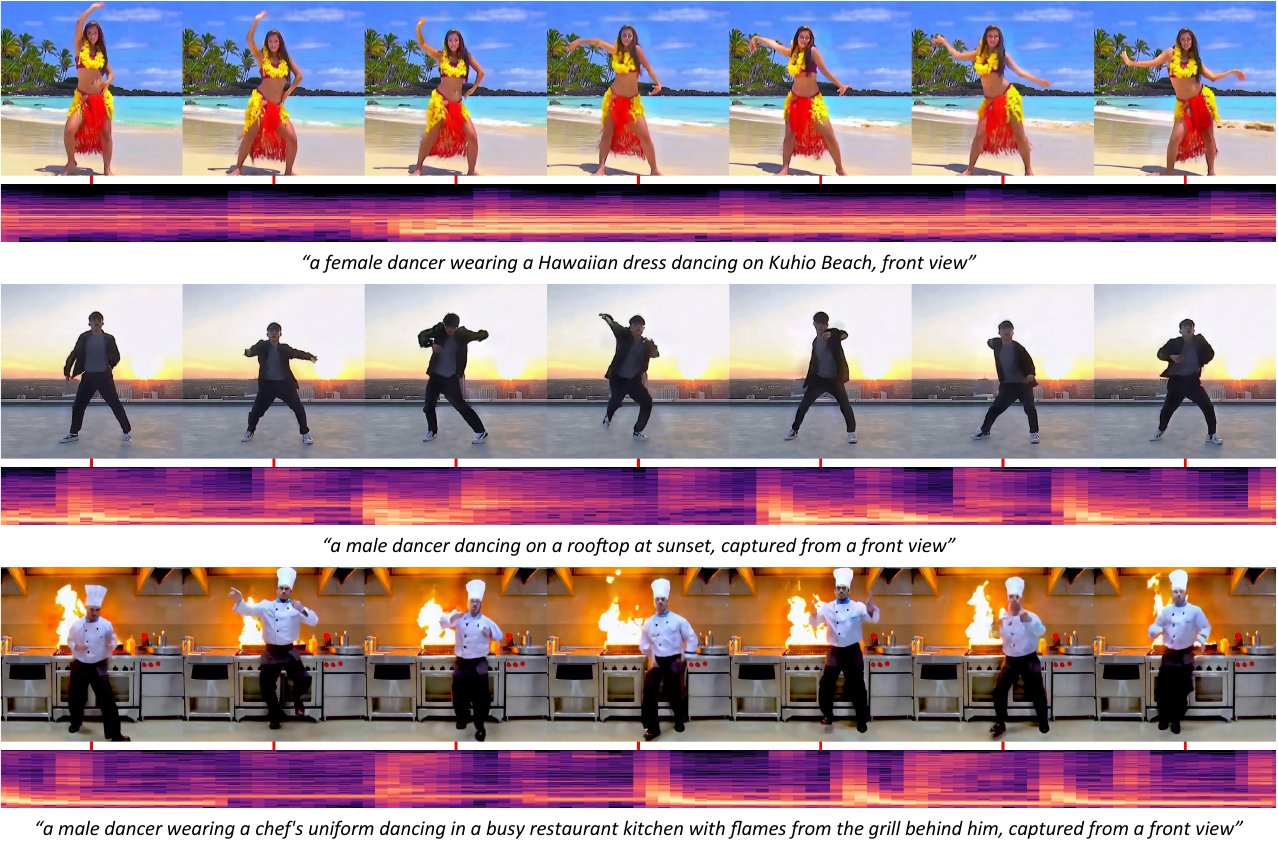}
\vspace{-10pt}
\caption{\textbf{MusicInfuser} adapts video diffusion models to music, making them listen and dance according to the music. This adaptation is done in a prior-preserving manner, enabling it to also accept style through the prompt while aligning the movement to the music.}
\label{fig:teaser}
\end{center}%
}]

\input{sec/0_abstract}

\input{sec/1_introduction}

\input{sec/2_related_work}

\input{sec/3_methods}

\input{sec/4_experiments}

\input{sec/5_conclusion}

{
    \small
    \bibliographystyle{ieeenat_fullname}
    \bibliography{main}
}

\input{sec/X_suppl}

\end{document}

%% file: preamble.tex
\usepackage{multirow}
\usepackage{amsthm}

\usepackage{wrapfig}

%% file: sec/0_abstract.tex
\begin{figure*}[t]
\center
\includegraphics[width=0.95\linewidth]{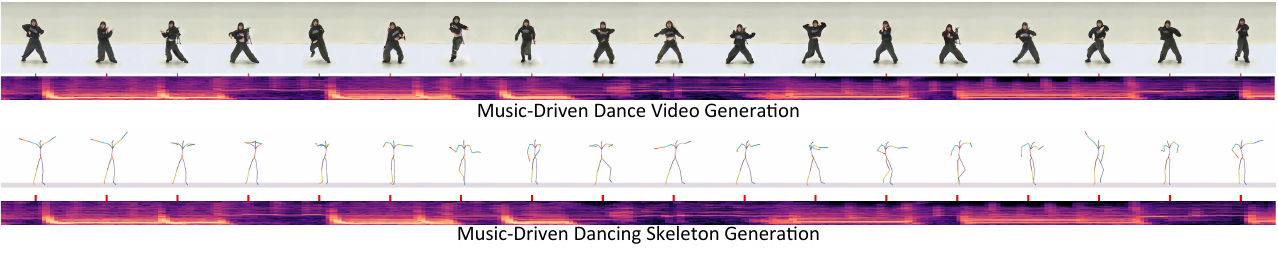}
\vspace{-10pt}
\caption{Motivational example. Skeletal motion generation~\cite{tseng2023edge} produces simplified movements lacking nuances such as backbone curvature, axial rotation, hand articulation, hair dynamics, and clothing motion, resulting in a more limited range of dance compared to video-based dance generation approaches (ours).}
\vspace{-10pt}
\label{fig:skeleton}
\end{figure*}

\begin{abstract}

We introduce MusicInfuser, an approach that aligns pre-trained text-to-video diffusion models to generate high-quality dance videos synchronized with specified music tracks. Rather than training a multimodal audio-video or audio-motion model from scratch, our method demonstrates how existing video diffusion models can be efficiently adapted to align with musical inputs. We propose a novel layer-wise adaptability criterion based on a guidance-inspired constructive influence function to select adaptable layers, significantly reducing training costs while preserving rich prior knowledge, even with limited, specialized datasets. Experiments show that MusicInfuser effectively bridges the gap between music and video, generating novel and diverse dance movements that respond dynamically to music. Furthermore, our framework generalizes well to unseen music tracks, longer video sequences, and unconventional subjects, outperforming baseline models in consistency and synchronization. All of this is achieved without requiring motion data, with training completed on a single GPU within a day.

\end{abstract}

%% file: sec/1_introduction.tex
\section{Introduction}
\label{sec:introduction}


Today's leading open-source video diffusion models often produce silent~\cite{genmo2024mochi} or speech-focused~\cite{kong2024hunyuanvideo,wan2025} videos. While it is possible to add music after the fact, it is difficult to generate motion that is properly synchronized with a specified music track. Alternatively, some research has begun to explore audio-video generation~\cite{ruan2023mm}. However, focusing on the specific application of {\em dance}, dance videos well-aligned with their music are far rarer than finding general, unconstrained videos, resulting in sub-optimal quality when training audio-video generative models from scratch.

In this paper, we introduce an approach to align {\em pre-trained} text-to-video models that have useful ingredients for dance. Our method, called {\em MusicInfuser}, generates output videos that are synchronized with the input music, with various components such as styles and appearances controllable via text prompts. We focus on synthesizing dance videos, i.e., generating realistic dancing figures that adjust and synchronize to the music, which poses several difficulties that require extensive knowledge about human motion and physics, music, and choreography.

Automatic dance generation must consider style, beat, and the inherently multimodal nature of dance, where multiple valid sequences can follow a given pose~\cite{lee2019dancing}. Computational approaches have drawn on choreographic principles~\cite{ye2020choreonet} and techniques ranging from graph-based methods~\cite{kim2003rhythmic,chen2021choreomaster} to deep neural networks~\cite{zhang2022music,zhuang2022music2dance,tseng2023edge}. However, traditional methods rely on motion capture~\cite{alexanderson2023listen} or reconstructed motions~\cite{li2021ai}, which are costly or prone to floating/jitter artifacts. In addition, skeletal representations are underparameterized for dance, lacking nuances such as backbone curvature, axial rotation, hand articulation, hair dynamics, and clothing motion (Fig.~\ref{fig:skeleton}).

MusicInfuser bypasses these limitations by adapting pre-trained text-to-video models~\cite{genmo2024mochi} with zero-initialized music-video modules injected into DiT blocks. This approach does not require motion capture or reconstruction, relying instead on existing dance videos for alignment. To address the scarcity of high-quality music-aligned dance datasets and reduce fine-tuning costs, we introduce a layer-wise adaptability criterion using a  {\em guidance-based constructive influence function}. This preserves pre-trained knowledge while establishing correlations between music and movement, allowing training on a GPU within a day.

MusicInfuser retains text-based control, enabling users to specify dance style, setting, and other aesthetics (Fig.~\ref{fig:teaser}) as well as the number of dancers (Fig.~\ref{fig:group}) while maintaining music synchronization. Our method generalizes to longer videos with unseen music and even to unseen subjects such as animals (Figs.~\ref{fig:animal},~\ref{fig:longer}). For evaluation, we introduce an automatic framework based on Video-LLMs~\cite{cheng2024videollama,xu2025qwen3} that jointly assesses video, audio, and language alignment, correlating well with human judgment.

Our experiments show that MusicInfuser successfully closes the gap between music and dance without intermediate motion data. By leveraging pre-trained video diffusion models through targeted adaptation, it produces high-quality, novel dance movements that respond naturally to musical rhythms, offering flexible dance video generation.

%% file: sec/2_related_work.tex
\begin{figure}[t]
\center
\includegraphics[width=0.95\linewidth]{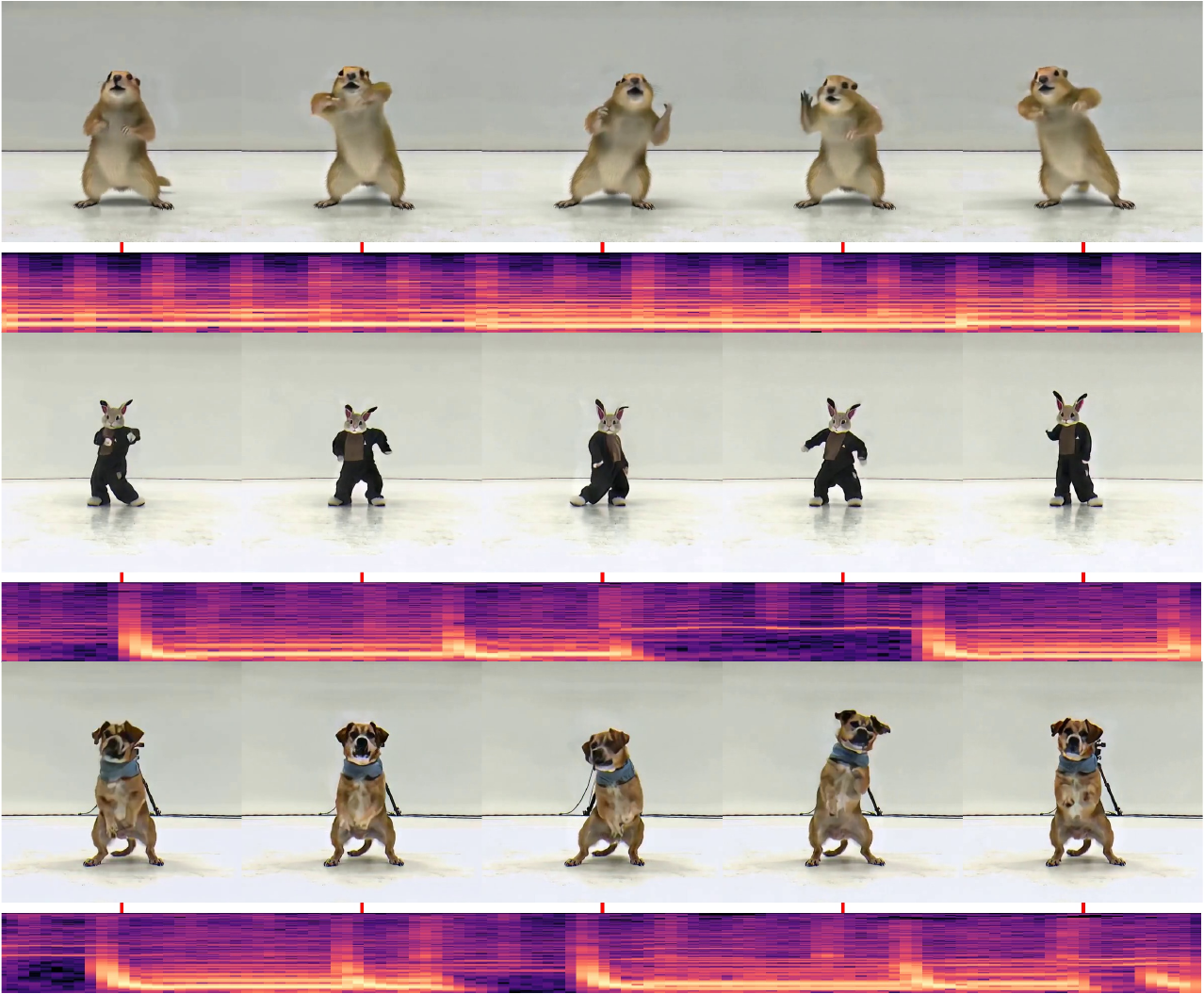}
\vspace{-10pt}
\caption{Using prompts such as ``a \{marmot, rabbit, dog (top to bottom rows)\} dancing ...," our method generalizes to unseen dancing subjects.}
\vspace{-10pt}
\label{fig:animal}
\end{figure}

\section{Related Work}
\label{sec:related-work}

\begin{figure}[t]
\center
\includegraphics[width=0.95\linewidth]{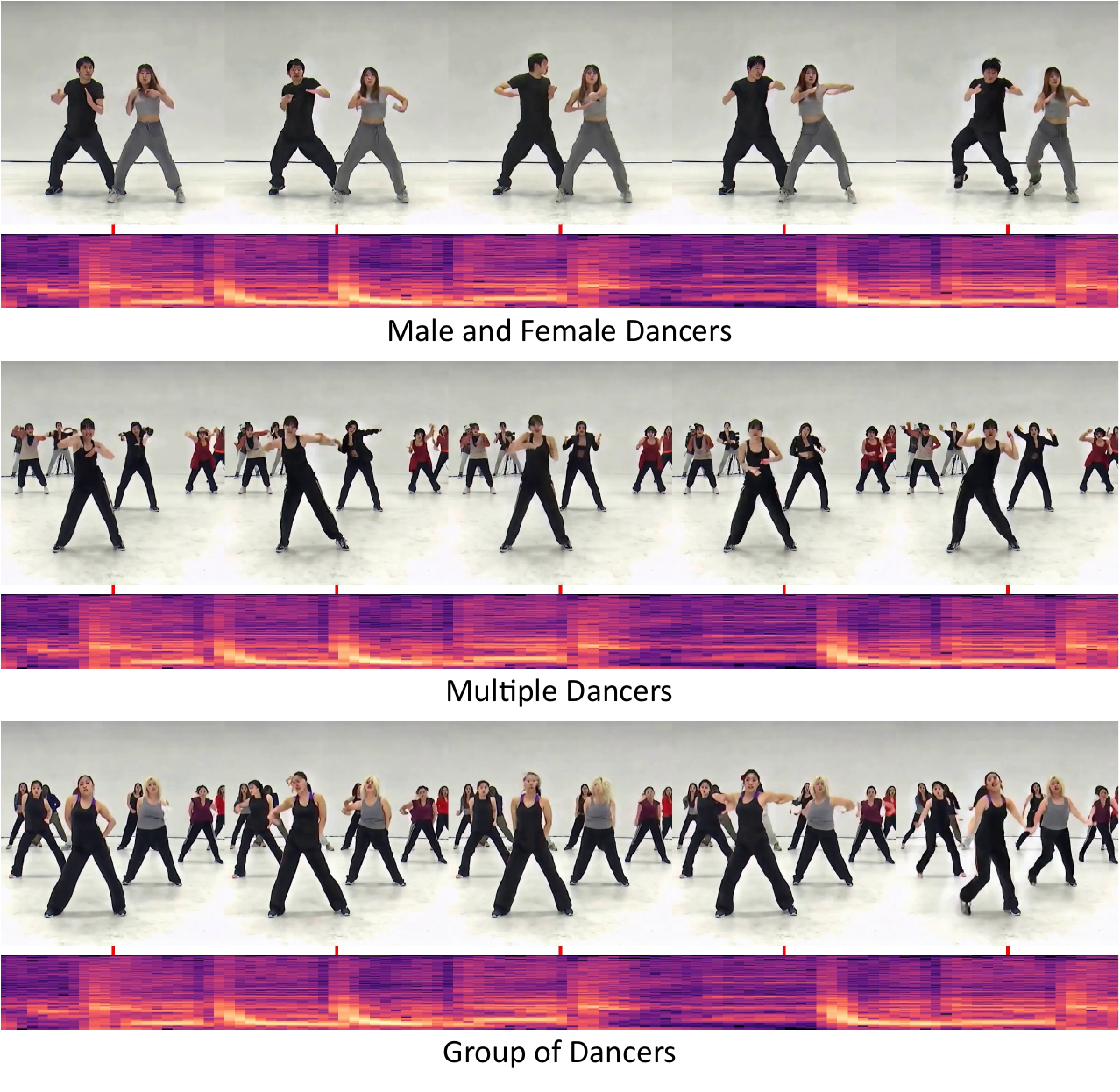}
\vspace{-10pt}
\caption{We can generate group dance videos aligned with music, based on the text.}
\vspace{-10pt}
\label{fig:group}
\end{figure}

\paragraph{Music-to-Dance Generation}
Early approaches mapped music primitives to dance elements using Hidden Markov Models~\cite{ofli2011learn2dance} and graph-based methods with movement transition graphs~\cite{kim2003rhythmic}. Later research integrated Gaussian processes~\cite{fukayama2014automated}, various neural networks~\cite{alemi2017groovenet,zhang2022music,zhuang2022music2dance}, and transformers~\cite{zhang2022music,li2021ai}. Traditional methods often produced beat-synchronized movements lacking contextual meaning or showing excessive repetition~\cite{aristidou2022rhythm}, with limited choreographic diversity~\cite{au2022choreograph}. Recent advances have shifted toward diffusion-based approaches~\cite{qi2023diffdance,alexanderson2023listen,tseng2023edge,qiu2024vimo,le2023controllable}, while supporting two- or multi-person dance~\cite{10264209,siyao2024duolando}. Unlike these skeleton-based methods, our framework directly synthesizes dance videos by adapting pre-trained text-to-video diffusion models to musical inputs. Without an intermediate representation, our method avoids rigid body parameterization, requires no motion capture or pose reconstruction, and eliminates post-processing to generate dance videos.

\vspace{-10pt}
\paragraph{Controllable Approaches}
Dance generation systems have evolved to incorporate multiple input modalities for richer choreographic control~\cite{chan2019everybody,liu2024dancegen,gong2023tm2d}, with text emerging as a powerful interface for its zero-shot capability and communicating choreographic ideas~\cite{liu2024dancegen}. Transformer-based approaches using Vector Quantized-Variational Autoencoders create discrete motion tokens processable alongside text~\cite{siyao2022bailando}, while systems now process both text and music inputs simultaneously~\cite{gong2023tm2d}. The MusicInfuser framework combines the flexibility of text-based interfaces with precise audio synchronization, allowing users to control stylistic and aesthetic elements of generated dance videos through prompts while ensuring movements remain aligned with musical features.

\vspace{-10pt}
\paragraph{Audio-to-Video Generation}

Another domain that is adjacent to our method is audio-driven video generation. Pioneering this domain, Sound2Sight~\cite{chatterjee2020sound2sight} introduced a deep variational encoder-decoder framework that predicts future frames by conditioning on both past frames and audio input. TATS~\cite{ge2022long} addressed audio-to-video generation challenges through a combination of time-agnostic VQGAN and time-sensitive transformer architectures. More recently, leveraging advances in diffusion models~\cite{ho2020denoising,song2020denoising}, joint audio-video generation methods like MM-Diffusion~\cite{ruan2023mm} have been developed, enabling bidirectional generation where either modality can condition the other.

\begin{figure*}[t]
\center
\includegraphics[width=0.95\linewidth]{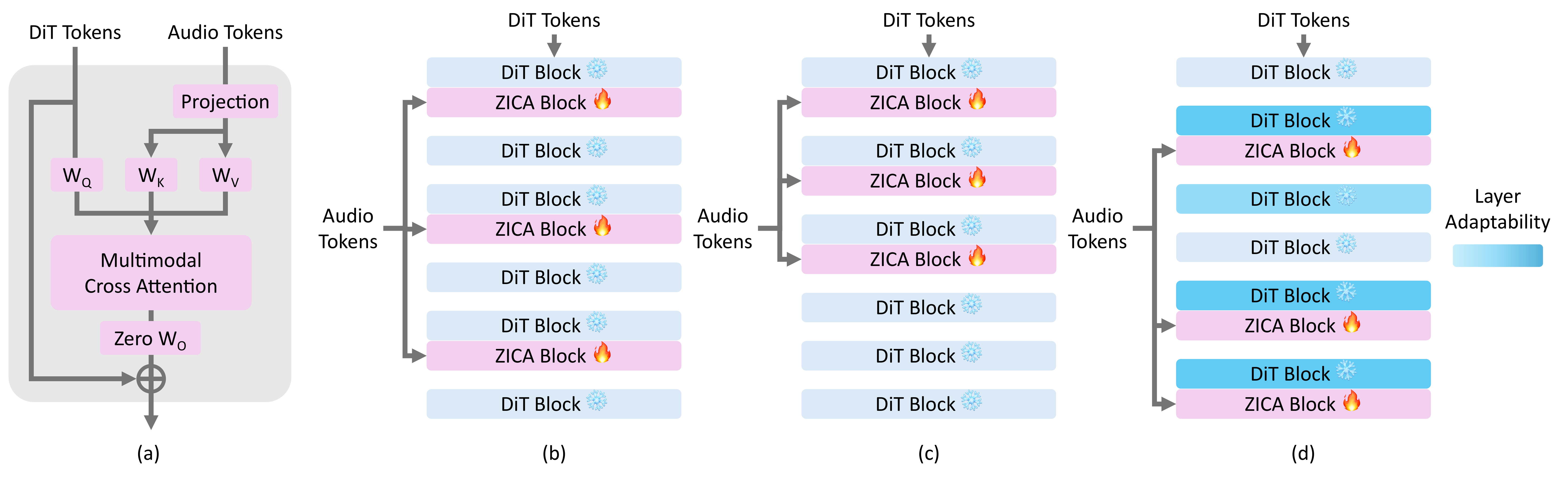}
\vspace{-10pt}
\caption{Zero-initialized cross-attention (ZICA) block. The output projection is initialized with a zero matrix, making the cross-attention block act as an identity function at the beginning. (b--d) illustrate several baseline layer selection strategies when the number of ZICA blocks is fewer than that of DiT blocks. (b) Attaching cross-attention blocks evenly across DiT blocks. (c) Attaching the blocks evenly to the earliest layers. (d) Attaching the blocks based on pre-computed layer adaptabilities (Sec.~\ref{sec:adaptability}). Table~\ref{tab:qwen-omni} shows the results.}
\vspace{-10pt}
\label{fig:zica-sch}
\end{figure*}

%% file: sec/3_methods.tex
\section{Preliminaries}
\label{sec:preliminaries}

\paragraph{Video Diffusion Models}
Diffusion models~\cite{ho2020denoising,song2020denoising,song2020score,ho2022video,blattmann2023stable,blattmann2023align,yu2023video} represent a family of generative techniques that restore data via iterative denoising steps. The goal is to generate samples from a video distribution $p(\mathbf{x})$. To this end, we can define a convoluted distribution of $p(\mathbf{x})$ and a Gaussian distribution with standard deviation $\sigma$, namely $p(\mathbf{x}, \sigma)$. In this paper, we follow~\citep{karras2022elucidating} to construct a compact formulation of diffusion models. The denoiser $D_\theta$ is optimized with the following L2 objective:
\begin{align}
\mathcal{L} = \mathbb{E}_{\mathbf{y} \sim p, \sigma \sim \Sigma_\text{train}, \mathbf{n} \sim \mathcal{N}(\mathbf{0}, \sigma^2\mathbf{I})}\|D_\theta(\mathbf{y}+\mathbf{n}; \sigma) - \mathbf{y}\|^2_2,
\end{align}
where $\Sigma_\text{train}$ denotes a noise distribution from which we sample noise during training, which is typically a uniform distribution.
To sample with the denoiser $D_\theta$, the ODE representing the change in the sample $\mathbf{x}$ with the change in $\sigma$ can be defined as:
\begin{align}
\frac{d\mathbf{x}}{d\sigma} = -\frac{D_\theta(\mathbf{x}; \sigma) - \mathbf{x}}{\sigma}.
\label{eq:denoiser}
\end{align}

\vspace{-10pt}
\paragraph{Text-Conditional Generation}
In a similar way, we can construct a conditional denoiser $D_\theta(\mathbf{x}|\mathbf{c}; \sigma)$ by training with a condition $\mathbf{c}$ paired with each $\mathbf{y}$ and replace the sampling process with a conditional denoiser.
To boost generated content quality and alignment with prompts, classifier-free guidance (CFG)~\cite{ho2022classifier} has become widely used. Applying CFG, the modified ODE then becomes the linear combination form:
\begin{align}
\frac{d\mathbf{x}}{d\sigma} = -\gamma_\text{cfg}\left[\frac{D_\theta(\mathbf{x}|\mathbf{c}; \sigma) - \mathbf{x}}{\sigma}\right] + (\gamma_\text{cfg} - 1)\left[\frac{D_\theta(\mathbf{x}; \sigma) - \mathbf{x}}{\sigma}\right].
\end{align}
In this formulation, $D_\theta(\mathbf{x}; \sigma)$ shares the same parameters as $D_\theta(\mathbf{x}|\mathbf{c}; \sigma)$ but is trained by randomly omitting conditional information during training, and parameter $\gamma_\text{cfg}$ denotes the guidance scale.

\section{MusicInfuser}
\label{sec:methods}

\paragraph{Text-to-video models already know how to dance.}

In contrast to previous multimodal dance generation methods, video diffusion models trained on massive and diverse video datasets have already internalized sophisticated representations of human motion, including intricate and expressive movements. They have implicitly learned choreographic patterns, style variations, and the general physics of a human body during their extensive training, providing a valuable foundation that can be leveraged for music-driven video generation.

Considering this, our goal is to align the models to musical input $\mathbf{a}$ with adaptation parameters $\phi$ to construct a final denoiser, $D_{\theta,\phi}(\mathbf{x}|\mathbf{c}, \mathbf{a}; \sigma)$, while preserving the pre-trained model's knowledge about dance. For the rest of this paper, we call the probability distribution characterized by the pre-trained text-to-video denoiser $D_\theta(\mathbf{x}|\mathbf{c}; \sigma)$ the prior, since it denotes a learned prior video distribution not conditioned on audio. Accordingly, our new continual optimization objective is as follows:
$$\mathcal{L} = \mathbb{E}_{\mathbf{(y, c, a)} \sim p_\textrm{mm}, \sigma \sim \Sigma_\text{train}, \mathbf{n} \sim \mathcal{N}(\mathbf{0}, \sigma^2\mathbf{I})}\|D_{\theta,\phi}(\mathbf{y} + \mathbf{n}|\mathbf{c}, \mathbf{a}; \sigma) - \mathbf{y}\|^2_2,$$
where $p_\textrm{mm}$ is a joint data distribution of video, caption, and audio.

Unfortunately, we recognize that specialized dancing datasets are notably scarcer than general video datasets used for pre-training and thus inevitably contain biases that compromise the prior model's generalization and denoising capabilities. Moreover, significant resources are required for fine-tuning text-to-video diffusion models. We address this challenge through a carefully balanced adaptation mechanism that preserves the rich prior while establishing robust correlations between musical features and dance movements with a significantly lower cost.

\begin{figure*}[t]
\center
\includegraphics[width=0.95\linewidth]{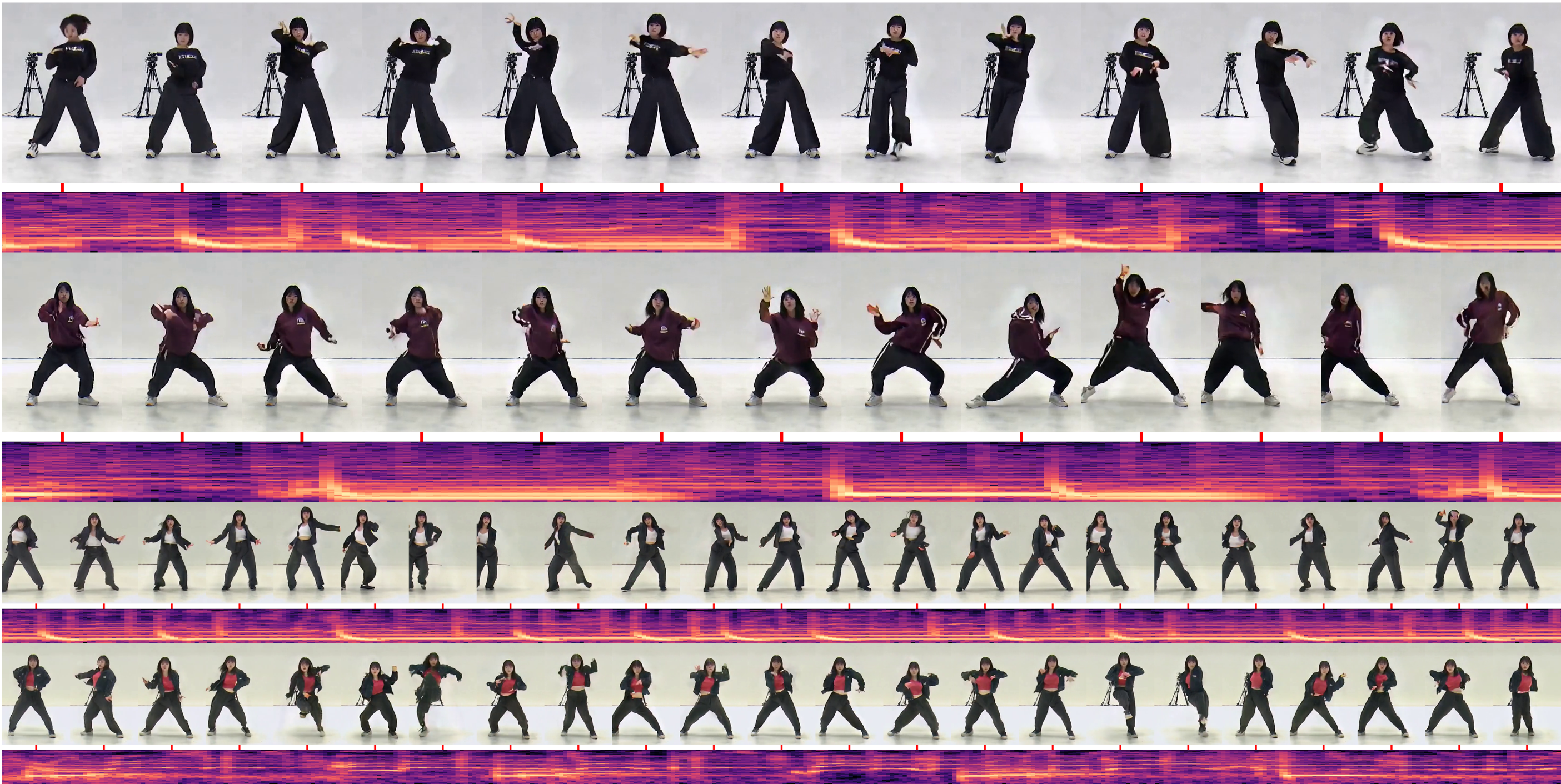}
\vspace{-10pt}
\caption{Generalization capabilities in terms of music length and type. MusicInfuser can generate multiple times longer dance videos that are multiple times longer than the videos used for training. For each row, we use synthetic in-the-wild music tracks with a keyword ``K-pop," a type of music not existing in AIST~\cite{tsuchida2019aist}, and use a prompt \textit{``a professional female dancer dancing K-pop ...."} This shows our method is highly generalizable, even extending to longer videos with an unseen category of the music. The beat and style alignment can be more clearly observed in the supplementary video.}
\label{fig:longer}
\vspace{-10pt}
\end{figure*}

\subsection{Measuring Layer Adaptability}
\label{sec:adaptability}

Cross-attention is effective for conditioning on auxiliary modalities~\cite{chen2024training,saharia2022photorealistic,hertz2022prompt}. However, applying cross-attention mechanisms to all layers of a model poses challenges: 1) it incurs substantial computational costs, and 2) it can degrade the denoising capabilities of pre-trained diffusion models (see ``All Layers" in Table~\ref{tab:qwen-omni}), especially in low-data regimes such as professional dancing. Consequently, identifying an optimal subset of layers for cross-attention adaptation is important to preserve both the denoising effectiveness and the generalization capability of the pre-trained model.

Unfortunately, finding this optimal combination through exhaustive search by fine-tuning every possible configuration is infeasible. For example, for a pre-trained model with 48 layers, the number of possible combinations for adapting only one-third of the layers with cross-attention is $\binom{48}{16} > 2 \times 10^{12}$. Two intuitive layer selection approaches, shown in (b) and (c) of Fig.~\ref{fig:zica-sch}, reduce computational costs. However, both methods fail to account for the behavior of the layers and compromise the model’s capabilities, specifically degrading video outputs (Table~\ref{tab:qwen-omni}).

To address this, we propose a principled, constructive metric for layer selection based on adaptability. Instead of measuring importance by performance degradation when a layer is removed, we measure each layer's \emph{positive influence} by using it as guidance during inference. This way, we can use existing evaluation metrics for videos~\cite{huang2024vbench} to measure and precompute the influence of each layer without the risk of out-of-distribution issues. Specifically, this criterion quantifies each layer's influence by performing guided sampling while leveraging the pre-trained model without the layer to provide guidance~\cite{hyung2024spatiotemporal,hong2025smoothed,ahn2024self}. The layer skip guidance can be formulated as the derivative of the implicit energy function~\cite{hyung2024spatiotemporal,hong2023improving}:
\begin{align}
\nabla_\mathbf{x}\mathcal{G}_l=\left({D_\theta^L(\mathbf{x}|\mathbf{c}; \sigma) - D_\theta^{L\backslash \{l\}}(\mathbf{x}|\mathbf{c}; \sigma)}\right)/{\sigma},
\end{align}
where $L$ represents the complete layer set, while $D_\theta^L$ and $D_\theta^{L\backslash \{l\}}$ denote the full-layer diffusion transformer denoiser and the variant skipping layer $l \in L$, respectively. Then, we define the improvement observed in the resulting videos as layer adaptability (see the supplementary material for details). Intuitively, layers that are more intrinsically connected to the structural and perceptual quality of video content exhibit greater performance when excluded and used for guidance than those primarily involved in local denoising. Our method thus identifies layers where modulation can effectively influence motion and structure, eliminating the need to train separate video models for each layer combination and avoiding significant deviations from the learned denoising manifold during adaptation.

\begin{figure*}[t]
\center
\includegraphics[width=0.95\linewidth]{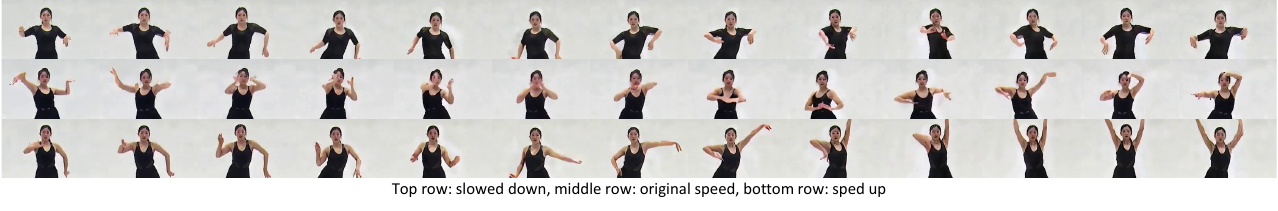}
\vspace{-10pt}
\caption{Speed control. The audio input is slowed down (top row) or sped up (bottom row) by factors of 0.75 and 1.25, respectively. This shows that speeding up audio generally results in sped-up movement. Note also the change in dynamics, as speeding up the audio increases the musical tone. More examples of audio speeding up and slowing down are included in the supplementary video.}
\label{fig:speed}
\vspace{-10pt}
\end{figure*}

\begin{table*}[t]
\centering
\resizebox{\textwidth}{!}{%
\begin{tabular}{l|cccccc|cccc|c}
\toprule
\multirow{2}{*}{\textbf{Model}} & \textbf{Style} & \textbf{Beat} & \textbf{Body} & \textbf{Movement} & \textbf{Choreography} & \textbf{Dance Quality} & \textbf{Imaging} & \textbf{Aesthetic} & \textbf{Overall} & \textbf{Video Quality} & \multirow{2}{*}{\textbf{Overall}} \\
& \textbf{Alignment} & \textbf{Alignment} & \textbf{Representation} & \textbf{Realism} & \textbf{Complexity} & \textbf{Average} & \textbf{Quality} & \textbf{Quality} & \textbf{Consistency} & \textbf{Average} & \\
\midrule
Ours & 8.95 & 9.54 & 10.00 & 7.36 & 5.25 & 8.22 & 7.08 & 7.01 & 9.96 & 8.02 & 8.14 \\
\midrule
All Layers & 8.37 & 9.02 & 9.55 & 7.02 & 5.35 & 7.86 & 6.00 & 7.17 & 9.95 & 7.71 & 7.80 \\
Evenly Distributed Layers & 8.15 & 9.01 & 9.95 & 6.59 & 5.08 & 7.76 & 5.93 & 6.61 & 9.66 & 7.40 & 7.62 \\
First Layers & 8.67 & 9.44 & 9.90 & 7.20 & 5.57 & 8.16 & 6.36 & 6.88 & 9.89 & 7.71 & 7.99 \\
Middle Layers & 9.05 & 9.39 & 9.05 & 6.67 & 5.56 & 7.94 & 5.84 & 6.80 & 9.78 & 7.47 & 7.77 \\
Last Layers & 8.60 & 9.34 & 9.80 & 6.91 & 5.45 & 8.02 & 6.03 & 6.70 & 9.81 & 7.51 & 7.83 \\
\midrule
Feature Addition & 8.60 & 9.37 & 9.90 & 6.92 & 5.41 & 8.04 & 6.29 & 6.90 & 9.94 & 7.71 & 7.92 \\
No Beta-Uniform Schedule & 8.93 & 9.42 & 9.40 & 6.67 & 5.61 & 8.01 & 6.20 & 6.99 & 9.93 & 7.71 & 7.89 \\
No In-the-Wild Data & 8.80 & 9.52 & 9.75 & 7.17 & 5.39 & 8.13 & 6.46 & 6.56 & 9.96 & 7.66 & 7.95 \\
No Cross Attention Zero Init. & 8.64 & 9.25 & 9.95 & 7.05 & 5.28 & 8.03 & 6.56 & 6.91 & 9.98 & 7.82 & 7.95 \\
\bottomrule
\end{tabular}
}
\vspace{-10pt}
\caption{Evaluation with Qwen3-Omni~\cite{xu2025qwen3} shows various baseline cross-attention adaptation strategies. Feature Addition refers to directly adding audio features, inspired by image conditioning in ControlNet~\cite{zhang2023adding}. Results using VideoLLaMA 2~\cite{cheng2024videollama} are provided in the supplementary material, which show a similar trend in layer selection.}
\vspace{-10pt}
\label{tab:qwen-omni}
\end{table*}

\subsection{Beta-Uniform Scheduling}
\label{sec:beta-uniform}

Diffusion models, including those using LoRA fine-tuning, typically employ a uniform distribution for noise sampling throughout training. For adapter training, we aim to preserve the denoising capability of the pre-trained model by initially focusing on low-noise levels and gradually learning the more substantial components over the course of training. To achieve this, we propose a Beta-Uniform scheduling strategy that evolves the training noise distribution $\Sigma_\text{train}$ from a Beta distribution concentrated on low noise levels to a uniform distribution.

The Beta distribution with parameters $\alpha=1$ and $\beta$ is formally defined by the probability density function:
\begin{align}
f(x; \alpha=1, \beta) = \frac{(1-x)^{\beta-1}}{B(1, \beta)}, \quad 0 \leq x \leq 1
\end{align}
where $B(\alpha, \beta)$ is the Beta function serving as a normalization constant. When $\beta > 1$, the distribution $\text{Beta}(1, \beta)$ concentrates probability mass near zero, which in our diffusion framework corresponds to sampling predominantly smaller noise scales. As $\beta$ decays toward 1, the distribution gradually flattens, approaching $\text{Uniform}(0, 1)$, i.e., $\lim_{\beta \to 1} f(x; 1, \beta) = 1$, for all $0 \leq x \leq 1$.

This causes a smooth transition from focusing on high-frequency components at lower noise levels to equally considering all frequencies. By first influencing the task-specific fine components of the dance and then the fundamental structure of dance movements, our approach preserves the pre-trained knowledge of general physics of human motion and produces more coherent dance sequences.

\subsection{Zero-Initialized Adaptation Modules}
\label{sec:zero-init}

To extend pre-trained diffusion transformers to new modalities while maintaining stable training, we introduce zero-initialized adaptation modules that start with zero parameters and gradually learn to influence the model. Specifically, we employ Zero-Initialized Cross-Attention (ZICA) for multimodal conditioning and Low-Rank Adaptors (LoRA) for domain and motion adaptation with the new modality.

Random initialization of cross-attention modules can bias predictions and destabilize continual training. ZICA addresses this by initializing the output projection to zero, so that cross-attention initially behaves as an identity mapping and gradually incorporates information from the conditioning modality. Let $\mathbf{A} \in \mathbb{R}^{N^A \times d}$ and $\mathbf{V} \in \mathbb{R}^{N^V \times d}$ denote projected audio and video tokens. The cross-attention with output projection is
\begin{align}
\mathbf{Z} = \mathbf{V} + \mathbf{W}_O \, \text{softmax}\left(\frac{\mathbf{V}\mathbf{W}_Q (\mathbf{A}\mathbf{W}_K)^\top}{\sqrt{d}}\right)\mathbf{A}\mathbf{W}_V,
\end{align}
where $\mathbf{W}_O$ is initialized to zero. The module initially acts as an identity mapping, and as $\mathbf{W}_O$ moves away from zero during training, it gradually integrates audio features.

Similarly, attention weights are adapted using LoRA~\cite{hu2022lora}, which decomposes updates into low-rank matrices initialized to zero. Conventional LoRA ranks (8--16) for image models are often insufficient for video transformers, which must capture temporal dependencies. We employ higher-rank configurations, since adapting temporal transformations generally requires higher rank. For example, full homography transformations requires increasing the rank by at least 8, and modeling complex human motion benefits from even higher ranks. LoRA, which is zero-initialized, parallels ZICA by gradually learning task-specific modifications from an initially neutral state, progressively adapting the diffusion network to the new domain and modality information.

\begin{table*}[t]
\centering
\footnotesize
\begin{tabular}{lccccccc}
\toprule
\multirow{2}{*}{\textbf{Model}} & \multirow{2}{*}{\textbf{Modality}} & \textbf{Style} & \textbf{Beat} & \textbf{Body} & \textbf{Movement} & \textbf{Choreography} & \textbf{Dance Quality} \\
& & \textbf{Alignment} & \textbf{Alignment} & \textbf{Representation} & \textbf{Realism} & \textbf{Complexity} & \textbf{Average} \\
\midrule
{\color{gray}{AIST Dataset} (GT)~\cite{tsuchida2019aist}} & {\color{gray}A+V} & {\color{gray}7.46} & {\color{gray}8.95} & {\color{gray}7.53} & {\color{gray}8.67} & {\color{gray}7.45} & {\color{gray}8.01} \\
MM-Diffusion~\cite{ruan2023mm} & A+V & 7.16 & 8.56 & 5.52 & 7.05 & 7.53 & 7.16 \\
Mochi~\cite{genmo2024mochi} & T+V & 7.20 & 8.34 & \textbf{7.47} & 7.68 & 7.82 & 7.70 \\
MusicInfuser (Ours) & T+A+V & \textbf{7.56} & \textbf{8.89} & 7.16 & \textbf{8.24} & \textbf{7.90} & \textbf{7.95} \\
\bottomrule
\end{tabular}
\vspace{-10pt}
\caption{Dance quality metrics comparing different models. A, V, and T denote audio, video, and text input modalities, respectively. For the models that have text input modality, we report an average of scores using a predefined benchmark of prompts.}
\vspace{-10pt}
\label{tab:dance-quality}
\end{table*}

\begin{table}[t]
\centering
\resizebox{\columnwidth}{!}{%
\begin{tabular}{lccccc}
\toprule
\multirow{2}{*}{\textbf{Model}} & \multirow{2}{*}{\textbf{Modality}} & \textbf{Imaging} & \textbf{Aesthetic} & \textbf{Overall} & \textbf{Video Quality} \\
 & & \textbf{Quality} & \textbf{Quality} & \textbf{Consistency} & \textbf{Average} \\
\midrule
{\color{gray}{AIST Dataset (GT)}~\cite{tsuchida2019aist}} & {\color{gray}A+V} & {\color{gray}9.76} & {\color{gray}8.17} & {\color{gray}9.77} & {\color{gray}9.23} \\
MM-Diffusion~\cite{ruan2023mm} & A+V & 8.94 & 6.52 & 8.38 & 7.94 \\
Mochi~\cite{genmo2024mochi} & T+V & 9.46 & \textbf{7.90} & 8.98 & 8.78 \\
MusicInfuser (Ours) & T+A+V & \textbf{9.60} & 7.87 & \textbf{9.39} & \textbf{8.95} \\
\bottomrule
\end{tabular}
}
\vspace{-10pt}
\caption{Video quality metrics comparing different models. For the models that have text input modality, we report an average of scores using a predefined benchmark of prompts.}
\vspace{-10pt}
\label{tab:video-quality}
\end{table}

\subsection{Utilizing In-the-Wild Data}
\label{sec:in-the-wild}
Training exclusively on datasets in highly constrained settings~\cite{tsuchida2019aist,li2021ai} can lead to reduced generalizability and model degradation when confronted with diverse real-world scenarios. Therefore, we use a mixture of in-the-wild data and the constrained datasets. These videos introduce diversity in terms of camera trajectories, lighting conditions, performance environments, and dance styles. The inclusion of in-the-wild data serves as regularization, preventing overfitting to specific dance patterns or environmental settings. Details are provided in the supplementary material.

\subsection{Prompt Diversification}
\label{sec:diversify-captions}

We use caption templates for constrained setting datasets that provide consistent and structured textual descriptions. These templates contain placeholders for key attributes such as dance style, setting, and movement quality, which are populated based on the specific characteristics of each video. For in-the-wild videos, which lack standardized descriptions, we use VideoChat2~\cite{li2023videochat} for generating captions. VideoChat2 analyzes the visual content and generates detailed captions that capture the contextual information present in these diverse video samples.

In addition, we randomly replace a small portion of detailed captions with basic, simple captions. This allows the adapter network to learn how to respond to music without relying on the text, effectively reducing the model's dependence on textual cues and encouraging it to develop stronger associations between musical features while still maintaining prompt adherence. This trade-off between style capture and music interpretation of the prompt is reflected in Table~\ref{tab:prompt-alignment}. The exact prompt templates used for diversification and replacement are provided in the supplementary material.

%% file: sec/4_experiments.tex
\section{Experiments}
\label{sec:experiments}

\subsection{Implementation Details}

\paragraph{Model Details}
We train our model on a single NVIDIA A100 GPU (except for experiments requiring more capacity) for 4,000 steps with a learning rate of $1\text{e}{-4}$, which takes roughly 20 hours to complete. Our LoRA uses rank 64, providing sufficient capacity to capture complex dance movements while maintaining parameter efficiency. For Beta-Uniform scheduling, we set the initial $\beta=3$ with exponential decay toward $\beta=1$. We use Mochi~\cite{genmo2024mochi} as our base model with a classifier-free guidance scale of $\gamma_\text{cfg}=6.0$ during inference and employ Wav2Vec 2.0~\cite{baevski2020wav2vec} as the audio encoder, while using a shallow MLP followed by downsampling to match the temporal dimension of the audio tokens for the audio projector.

\vspace{-10pt}
\paragraph{Dataset}
The AIST dataset~\cite{tsuchida2019aist} includes 13,940 videos with 60 musical pieces, 10 dance genres, and 35 dancers. We extract 2,378 clips and divide the training and test sets with non-overlapping music tracks, following AIST++~\cite{li2021ai}. We randomly sample approximately 2.5-second clips from full sequences for training. As mentioned in Sec.~\ref{sec:in-the-wild}, we supplement AIST with 15,799 in-the-wild dance video clips from 4 YouTube playlists containing over 3.7k videos across various dance styles and settings. These clips are mixed with the AIST data at a 1:1 ratio during training, creating a balanced dataset that combines AIST's controlled studio environment with diverse real-world performances.

\vspace{-10pt}
\paragraph{Quantitative Metrics}

Evaluating generated content automatically presents challenges. Inspired by VBench's use of Visual-Language Models for text-to-video assessment~\cite{li2023videochat}, we propose a novel metric using VideoLLaMA 2~\cite{cheng2024videollama} and Qwen3-Omni~\cite{xu2025qwen3}, which process both video and audio inputs. We formulate targeted queries to assess three components: dance quality (style alignment, beat alignment, body representation, movement realism, choreography complexity), video quality (imaging quality, aesthetic quality, overall consistency), and prompt alignment (style capture, creative interpretation, satisfaction). See the supplementary material for exact prompts and methods. Tables~\ref{tab:dance-quality} and \ref{tab:video-quality} also display results on AIST test data, where the ground truth data outperforms generated content in metrics like beat alignment and movement realism. This validates that our metric correctly assigns higher scores to ground truth data, which should represent the upper bound for these metrics, and demonstrates its reliability alongside correlation with human evaluation results shown in Fig.~\ref{fig:human}.

\subsection{Experimental Results}

\paragraph{Music- and Text-Driven Dance Video Generation}
Fig.~\ref{fig:teaser} showcases the model's ability to combine textual control with musical synchronization. The generated videos successfully incorporate scene contexts (restaurant kitchen, beach at sunset) and dancer attributes (wearing a leather jacket, chef's uniform) as specified in the prompts, while simultaneously aligning the choreographic style with the musical input. Figs.~\ref{fig:animal} and~\ref{fig:group} demonstrates that our model is capable of generating unseen subjects or rare settings.

\vspace{-10pt}
\paragraph{Music Responsiveness}
In Fig.~\ref{fig:speed}, we show how MusicInfuser generates dance videos, including the movement and outfit of the dancer, based on the music condition, while keeping the prompt fixed. Additionally, we demonstrate the model's responsiveness to musical features through experiments with tempo modification. By accelerating the music track by 1.25 times or decelerating it by 0.75 times, the generated dance movements appropriately adjust pace while maintaining similar choreographic style, as shown in Fig.~\ref{fig:speed}. Furthermore, acceleration and deceleration also result in changes in tone, which affect the dynamicity of the dance generated by our model. This shows that our model successfully captures the relationship between musical tempo and dance movement dynamicity, a critical aspect of dance-music synchronization.

\begin{table}[t]
\centering
\resizebox{\columnwidth}{!}{%
\begin{tabular}{lcccc}
\toprule
\multirow{2}{*}{\textbf{Model}} & \textbf{Style} & \textbf{Creative} & \textbf{Overall} & \textbf{Prompt Align} \\
& \textbf{Capture} & \textbf{Interpretation} & \textbf{Satisfaction} & \textbf{Average} \\
\hline
Mochi~\cite{genmo2024mochi} & \textbf{7.98} & 9.04 & 9.55 & 8.86 \\
MusicInfuser (Ours) & 7.80 & \textbf{9.27} & \textbf{9.80} & \textbf{8.96} \\
\hline
No in-the-Wild Data & 6.80 & 8.69 & 8.40 & 7.96 \\
Base Prompt 0\% & 7.45 & 8.85 & 9.43 & 8.58 \\
Base Prompt 100\% & 7.33 & 9.06 & 9.36 & 8.58 \\
\bottomrule
\end{tabular}
}
\vspace{-10pt}
\caption{Prompt alignment metrics comparing different models.}
\vspace{-10pt}
\label{tab:prompt-alignment}
\end{table}

\begin{figure}[t]
\center
\includegraphics[width=0.95\linewidth]{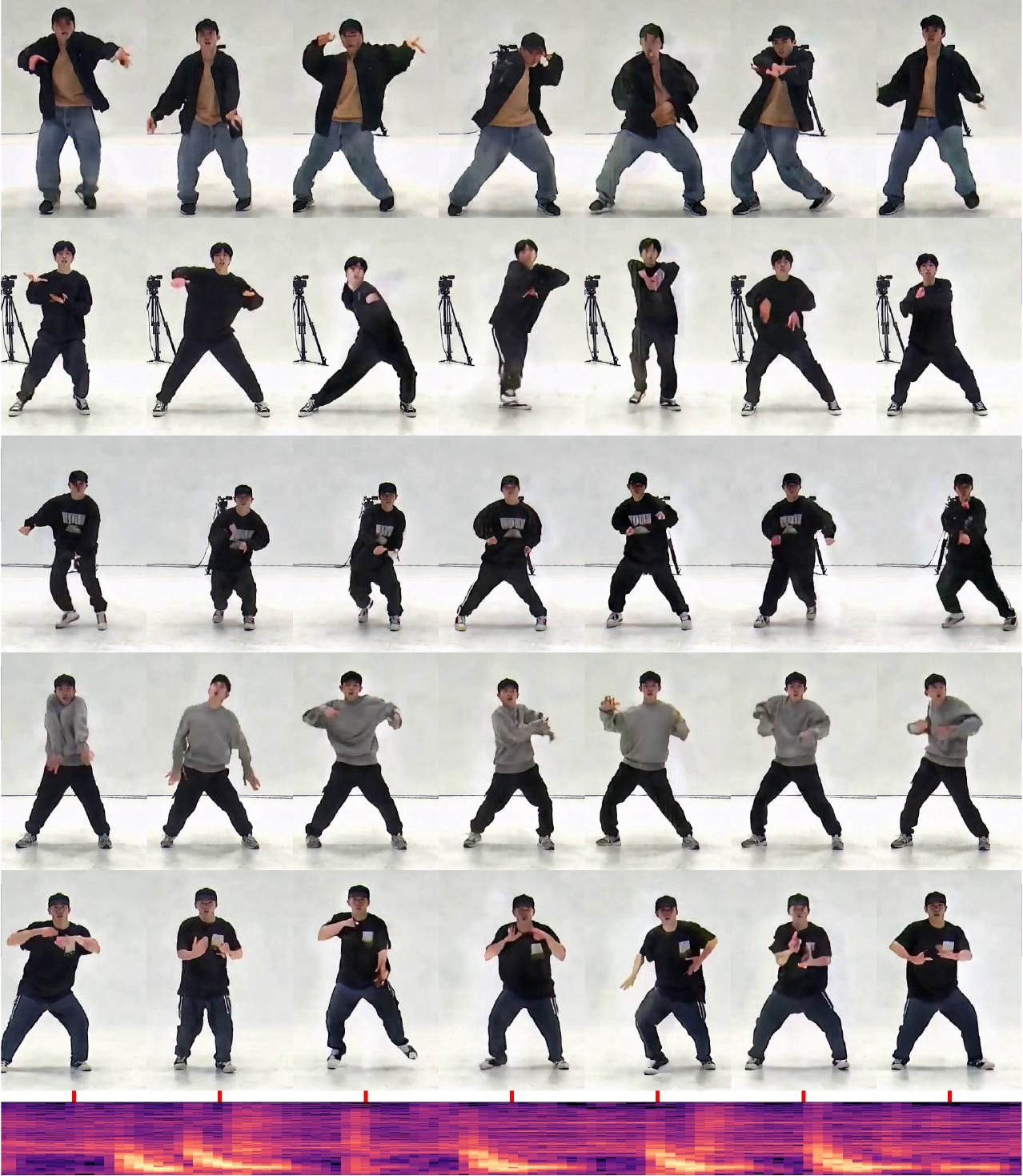}
\vspace{-10pt}
\caption{By changing the seed, our method can produce diverse results given the same music and text. The generated choreography of each dance is different from each other. We use the fixed prompt ``a professional dancer dancing ...."}
\vspace{-10pt}
\label{fig:diversity}
\end{figure}

\vspace{-10pt}
\paragraph{Generalization to In-the-Wild Music and Longer Videos}
To evaluate generalization beyond the AIST music distribution, we test our model with music tracks generated by SUNO AI. Fig.~\ref{fig:longer} shows successful generation for these unseen music categories, confirming the model's ability to map novel audio patterns to appropriate dance movements. In addition, Fig.~\ref{fig:longer} shows longer video generation results with the same setting but with multiple times as many frames as the videos we used for training, up to 9 seconds.

\vspace{-10pt}
\paragraph{Baseline Comparison}

We present several baselines for our layer adaptation in Table~\ref{tab:qwen-omni}. Adapting the layers with cross-attention that we selected based on the layer adaptability criterion in Sec.~\ref{sec:adaptability} significantly outperforms the strategy of selecting evenly distributed layers, first layers, middle layers, and last layers, and even outperforms adapting all layers in the video diffusion model. This demonstrates that our positive influence function for layer selection is crucial for high-performance adaptation.

Tables~\ref{tab:dance-quality}--\ref{tab:video-quality} present our quantitative results compared against prior work~\cite{ruan2023mm,genmo2024mochi}. For dance quality (Table~\ref{tab:dance-quality}), our method outperforms previous approaches in style alignment, beat alignment, movement realism, and choreography complexity, while maintaining competitive scores across other metrics. Table~\ref{tab:video-quality} demonstrates our superiority in video quality metrics, particularly in imaging quality and overall consistency compared to MM-Diffusion~\cite{ruan2023mm} and Mochi~\cite{genmo2024mochi}. In Table~\ref{tab:prompt-alignment}, MusicInfuser shows improved creative interpretation and overall satisfaction over the baseline Mochi model. For qualitative comparisons with prior work, we refer readers to the supplementary material.

\vspace{-10pt}
\paragraph{Human Evaluation}

\begin{figure}[t]
\center
\includegraphics[width=0.9\linewidth]{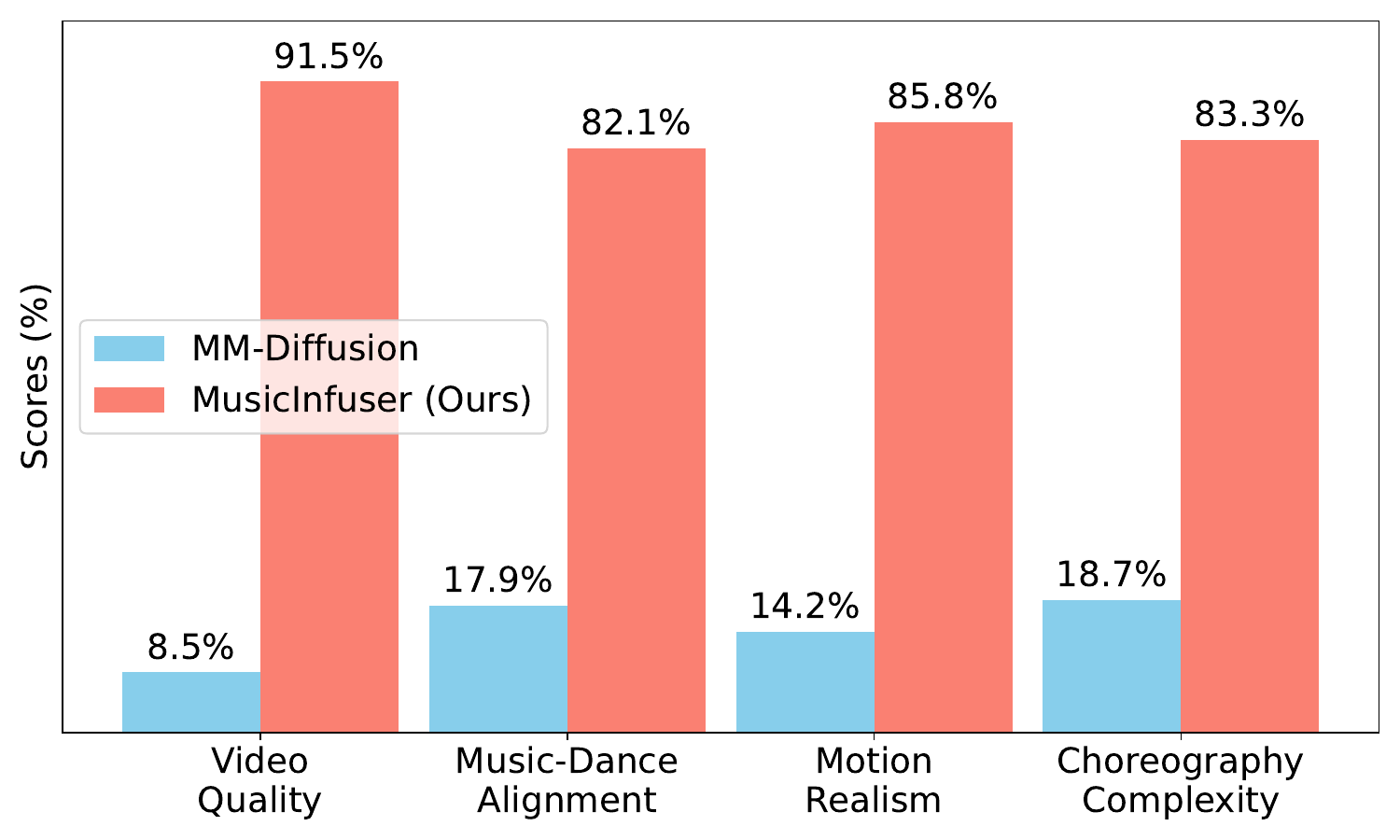}
\vspace{-10pt}
\caption{Human evaluation.}
\vspace{-10pt}
\label{fig:human}
\end{figure}

We conduct human evaluation to validate MusicInfuser's performance and examine the correlation between Video-LLM-based quantitative assessment (Tables \ref{tab:dance-quality} and \ref{tab:video-quality}) and human judgments. Fig.~\ref{fig:human} presents the results of our human evaluation study, where we assess generated videos across multiple dimensions including video quality, music-dance alignment, motion realism, and choreography complexity. The human evaluation demonstrates that our approach consistently outperforms previous work, with evaluators particularly noting improvements in video quality and movement naturalness. The details of the human evaluation are provided in the supplementary material.

\vspace{-10pt}
\paragraph{Ablation Studies}
In Table~\ref{tab:qwen-omni}, we evaluate the contribution of the components in our framework. Using Beta-Uniform scheduling improves body representation and movement realism. The naive feature addition baseline, where instead of using the ZICA adapter we simply spatially expand the audio feature and add it to the corresponding frame, similar to ControlNet~\cite{zhang2023adding}, performs worse than our approach in most metrics, confirming the effectiveness of our ZICA strategy. Not zero-initializing the cross-attention layers~\cite{ye2023ip} results in a remarkable drop in the video quality metric. Additionally, in Table~\ref{tab:prompt-alignment}, we show the trade-off between style capture and creative interpretation of the prompt depending on the base prompt ratio, meaning how frequently we replaced the prompt with the basic prompt. More ablation studies and analysis are in the supplementary material.

\vspace{-10pt}
\paragraph{Diversity of Results}
By varying the random seed while keeping the prompt and music constant, our model generates diverse choreographies, as shown in Fig.~\ref{fig:diversity}, demonstrating that it does not simply memorize routines for particular tracks but is capable of generating diverse dance sequences.

%% file: sec/5_conclusion.tex
\section{Conclusion}
\label{sec:conclusion}

In this paper, we present MusicInfuser, a novel approach for generating dance videos synchronized with music by leveraging the rich choreographic knowledge embedded in pre-trained text-to-video diffusion models. Through our adaptation architecture and strategies, MusicInfuser enables synchronized dance movements with musical inputs while preserving text-based control over style and scene elements. It achieves this without requiring expensive motion capture data, generalizes to novel music tracks and subjects, and supports the generation of diverse choreographies.

\section*{Acknowledgments}
We thank Xiaojuan Wang and Jingwei Ma for their valuable feedback. This work was supported by the UW Reality Lab and Google.

%% file: sec/X_suppl.tex
\clearpage
\appendix

\onecolumn
\begin{center}
\Large
\textbf{\thetitle}\\
\vspace{0.5em}Supplementary Material \\
\vspace{1.0em}
\end{center}

\begin{figure*}[h]
\center
\includegraphics[width=1.0\linewidth]{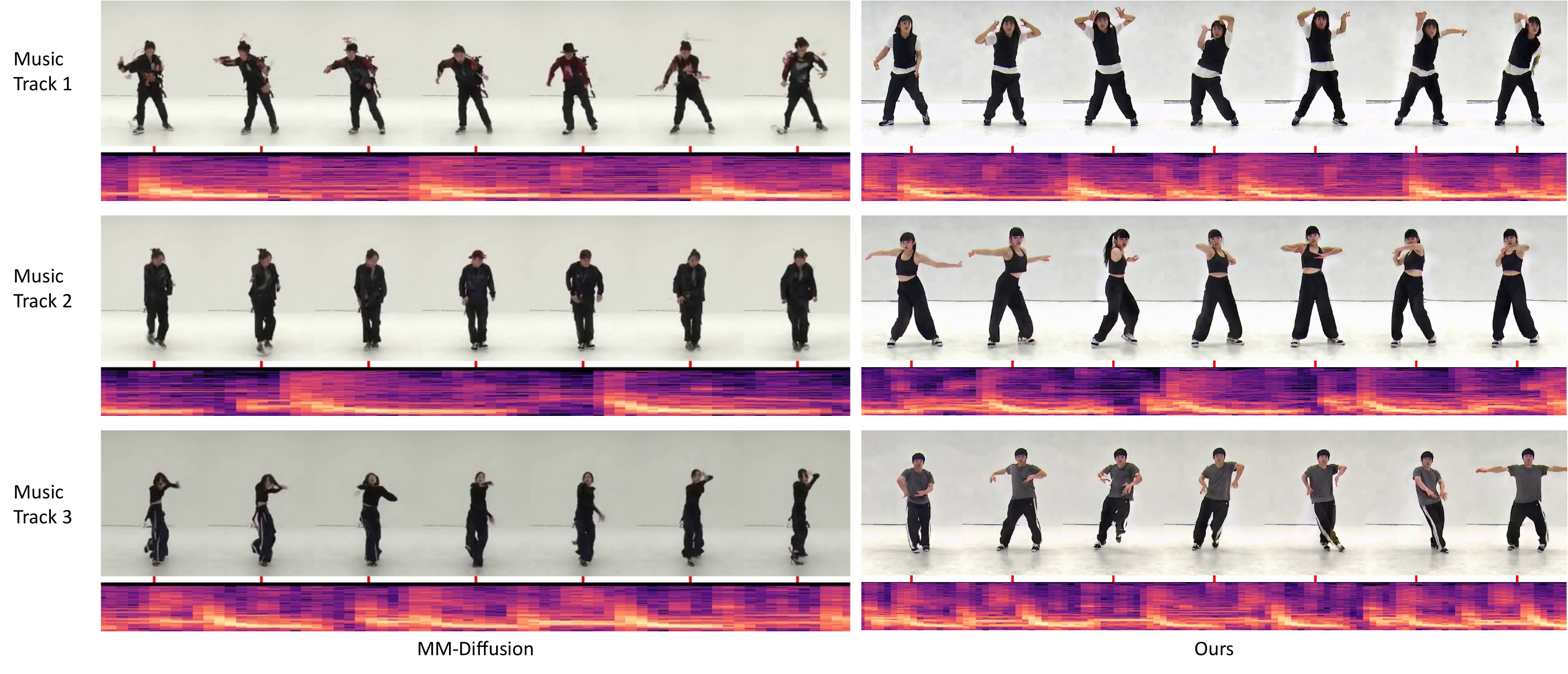}
\caption{Comparison of audio-driven generation with MM-Diffusion~\cite{ruan2023mm}. Our method produces fewer artifacts (shown in the first and third rows), while generating more realistic dance videos with more natural movements (first row) and more dynamic motion (second and third rows). Note that we use the same music track for each row, and the spectrogram is stretched for MM-Diffusion since we generate longer videos. For our method, we use the fixed caption ``a professional dancer dancing ..." across all music tracks.}
\label{fig:mm-diffusion}
\end{figure*}

\begin{figure*}[h]
\center
\includegraphics[width=1.0\linewidth]{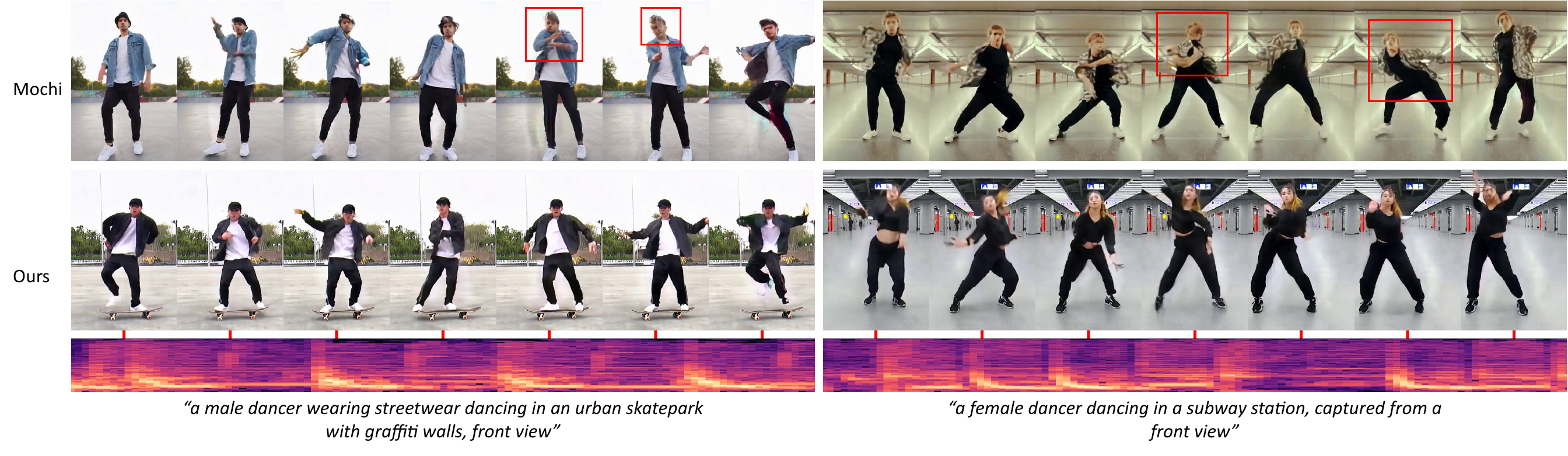}
\caption{MusicInfuser infuses listening capability into the text-to-video model (Mochi~\cite{genmo2024mochi}), while preserving prompt adherence and improving overall consistency and realism.
}
\label{fig:mochi-comparison}
\end{figure*}

\section{Video Results}

We present the flattened video results along the time axis and the corresponding spectrograms in the main paper. However, our frame sampling rate does not exceed the Nyquist frequency for the general musical beat, causing the movement to appear slower. Therefore, we encourage readers to view the supplementary video.

\begin{figure}[t]
\centering
\includegraphics[width=\linewidth]{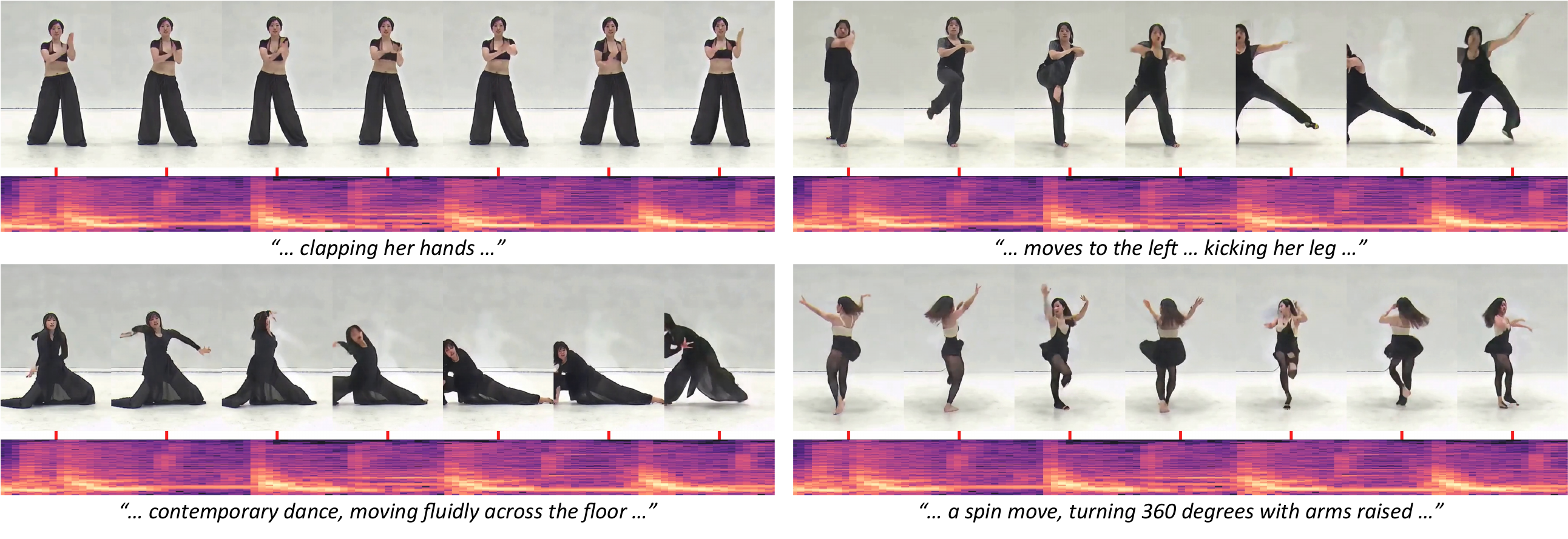}
\caption{Text-based dance control.}
\label{fig:text-control}
\end{figure}

\begin{table*}[t]
\centering
\resizebox{\textwidth}{!}{%
\begin{tabular}{l|cccccc|cccc|c}
\toprule
\multirow{2}{*}{\textbf{Layer Selection Strategy}} & \textbf{Style} & \textbf{Beat} & \textbf{Body} & \textbf{Movement} & \textbf{Choreography} & \textbf{Dance Quality} & \textbf{Imaging} & \textbf{Aesthetic} & \textbf{Overall} & \textbf{Video Quality} & \multirow{2}{*}{\textbf{Overall}} \\
& \textbf{Alignment} & \textbf{Alignment} & \textbf{Representation} & \textbf{Realism} & \textbf{Complexity} & \textbf{Average} & \textbf{Quality} & \textbf{Quality} & \textbf{Consistency} & \textbf{Average} & \\
\hline
Layer Adaptability & 7.56 & 8.89 & 7.16 & 8.24 & 7.90 & 7.95 & 9.60 & 7.87 & 9.39 & 8.95 & 8.33 \\
Evenly Distributed Layers & 7.31 & 8.81 & 7.28 & 7.70 & 7.96 & 7.81 & 9.33 & 7.78 & 9.04 & 8.72 & 8.15 \\
First Layers & 7.25 & 8.82 & 6.86 & 7.37 & 8.05 & 7.67 & 9.66 & 7.91 & 9.27 & 8.95 & 8.15 \\
Middle Layers & 7.91 & 8.87 & 6.74 & 7.83 & 7.98 & 7.86 & 9.21 & 7.97 & 9.20 & 8.79 & 8.21 \\
Last Layers & 7.52 & 8.81 & 7.01 & 7.47 & 8.00 & 7.76 & 9.45 & 7.73 & 9.14 & 8.77 & 8.14 \\
All Layers & 7.49 & 8.53 & 6.72 & 8.16 & 7.85 & 7.75 & 9.33 & 7.99 & 9.11 & 8.81 & 8.15 \\
\bottomrule
\end{tabular}
}
\caption{Evaluation of layer selection strategies using VideoLLaMA 2~\cite{cheng2024videollama}.}
\label{tab:layer}
\end{table*}

\begin{table*}[t]
\centering
\resizebox{\textwidth}{!}{%
\begin{tabular}{l|cccccc|cccc|c}
\toprule
\multirow{2}{*}{\textbf{Model}} & \textbf{Style} & \textbf{Beat} & \textbf{Body} & \textbf{Movement} & \textbf{Choreography} & \textbf{Dance Quality} & \textbf{Imaging} & \textbf{Aesthetic} & \textbf{Overall} & \textbf{Video Quality} & \multirow{2}{*}{\textbf{Overall}} \\
& \textbf{Alignment} & \textbf{Alignment} & \textbf{Representation} & \textbf{Realism} & \textbf{Complexity} & \textbf{Average} & \textbf{Quality} & \textbf{Quality} & \textbf{Consistency} & \textbf{Average} & \\
\hline
Full (Ours) & 7.56 & 8.89 & 7.16 & 8.24 & 7.90 & 7.95 & 9.60 & 7.87 & 9.39 & 8.95 & 8.33 \\
No ZICA Layer Selection & 7.31 & 8.81 & 7.28 & 7.70 & 7.96 & 7.81 & 9.33 & 7.78 & 9.04 & 8.72 & 8.15 \\
No Higher Rank & 7.37 & 8.76 & 6.86 & 7.75 & 7.98 & 7.74 & 9.55 & 7.94 & 9.49 & 8.99 & 8.21 \\
No LoRA & 7.48 & 8.62 & 7.02 & 7.53 & 7.95 & 7.72 & 9.43 & 8.08 & 9.36 & 8.96 & 8.18 \\
No Beta-Uniform Schedule & 8.04 & 9.07 & 6.35 & 7.88 & 7.91 & 7.85 & 9.17 & 7.85 & 9.37 & 8.80 & 8.21 \\
Feature Addition & 7.62 & 8.90 & 6.78 & 7.97 & 7.88 & 7.83 & 9.44 & 7.88 & 9.31 & 8.88 & 8.22 \\
\bottomrule
\end{tabular}
}
\caption{Ablation study. Feature addition denotes that we spatially expand the audio feature and add it to the corresponding frame. We use VideoLLaMA 2~\cite{cheng2024videollama} for the evaluation.}
\label{tab:abl}
\end{table*}

\section{Text-Driven Choreography}

To demonstrate text-based control of dance sequences, we conducted additional experiments in which text prompts guide the dance dynamics (clapping, kicking, spinning, etc.). The results are shown in Fig.~\ref{fig:text-control}. This demonstrates that MusicInfuser can also generate diverse music-synchronized videos that follow the motion directions specified in the text prompt.

\section{Dance Difficulty Control}

\begin{figure}[ht]
\center
\includegraphics[width=0.6\linewidth]{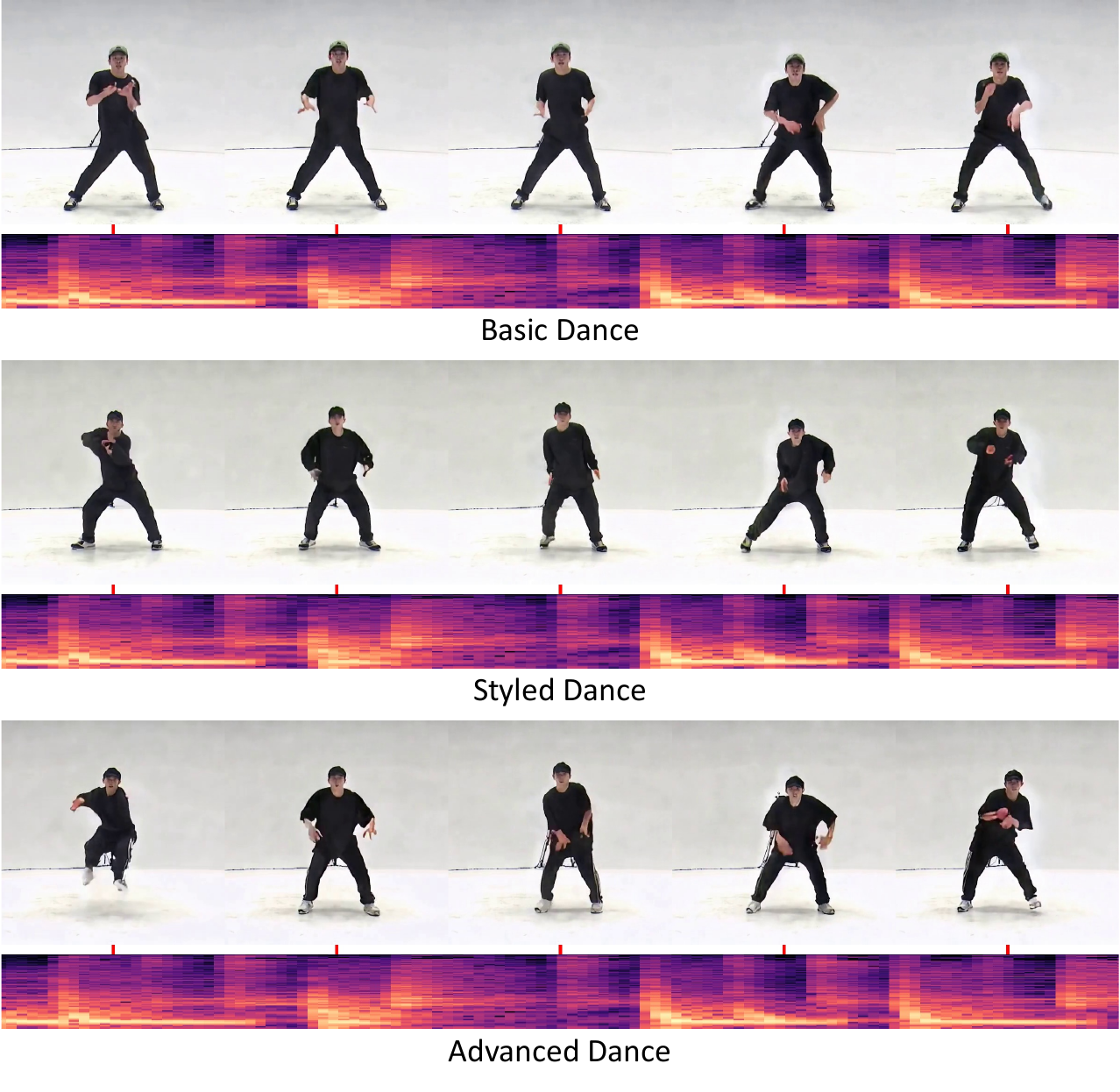}
\caption{Changes in the complexity of choreography.}
\label{fig:difficulty}
\end{figure}

We demonstrate difficulty control of the choreography in Fig.~\ref{fig:difficulty}, which is achieved using the same seed and music but with prompts of varying specificity. For basic dance, we use the general prompt ``a professional dancing in a studio with a white backdrop.'' For styled dance, we additionally specify the dance genre but use ``basic dance setting,'' and for advanced, we change it to ``advanced dance setting.''

\begin{figure}[ht]
\center
\includegraphics[width=1.0\linewidth]{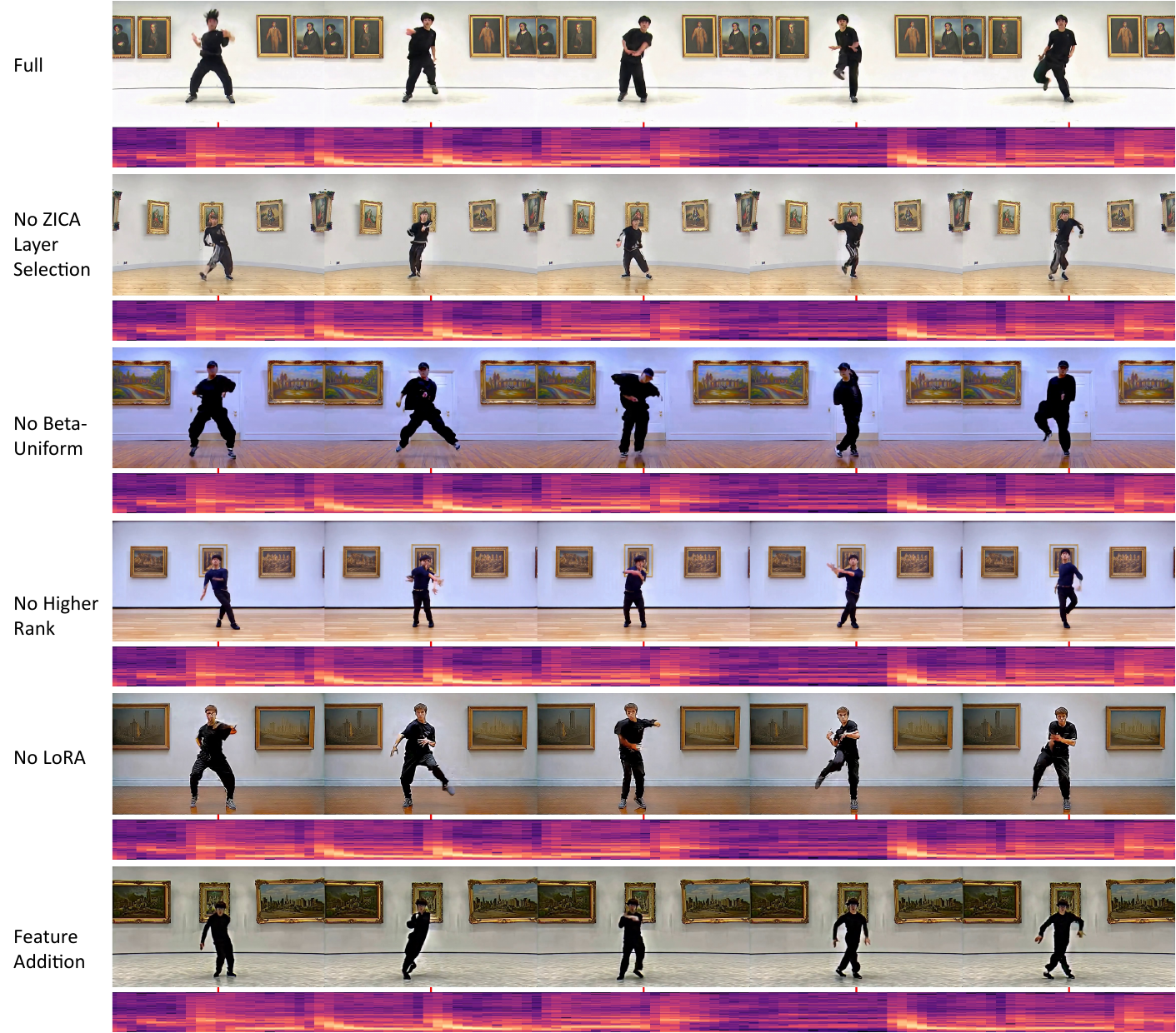}
\caption{Ablation study. The prompt is set to ``a male dancer dancing in an art gallery with some paintings, captured from a front view''. The seed and music are set the same across all methods.}
\label{fig:abl-1}
\end{figure}

\begin{figure}[ht]
\center
\includegraphics[width=1.0\linewidth]{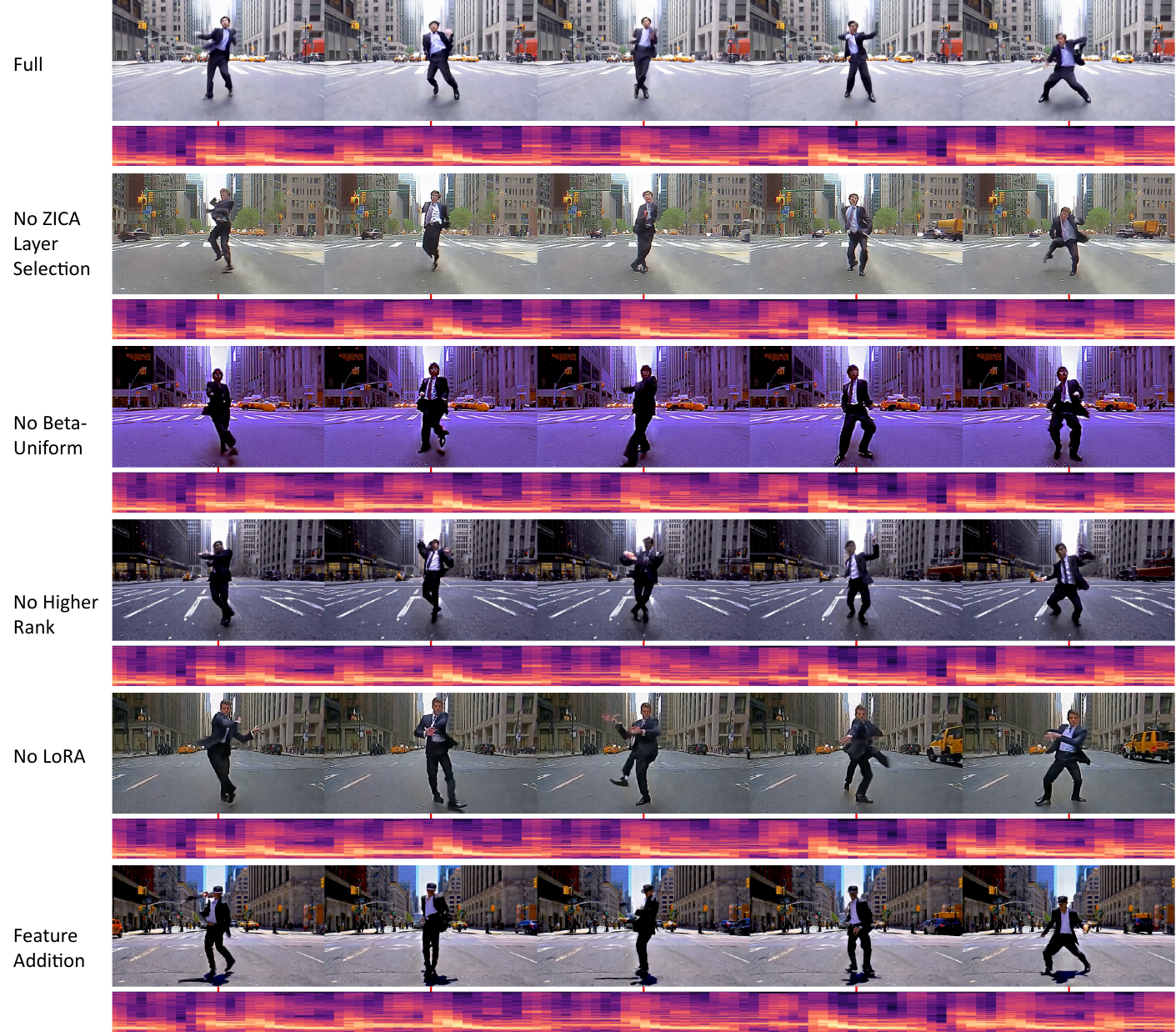}
\caption{Ablation study. The prompt is set to ``a male dancer wearing a suit dancing in the middle of a New York City, captured from a front view''. The seed and music are set the same across all methods.}
\label{fig:abl-2}
\end{figure}

\section{Human Evaluation Protocol}

For each test music track~\cite{li2021ai}, we conducted fully anonymized A/B testing. We asked 33 participants to evaluate video quality, music-dance alignment, motion realism, and choreography complexity. The following are examples of the questionnaire items:
\begin{enumerate}
    \item Which video has higher visual quality?
    \item Which video's dance aligns better with the music?
    \item Which video's motion is more realistic?
    \item Which video's dance is more complex?
\end{enumerate}

\section{Limitations}

\begin{figure}[ht]
\center
\includegraphics[width=0.6\linewidth]{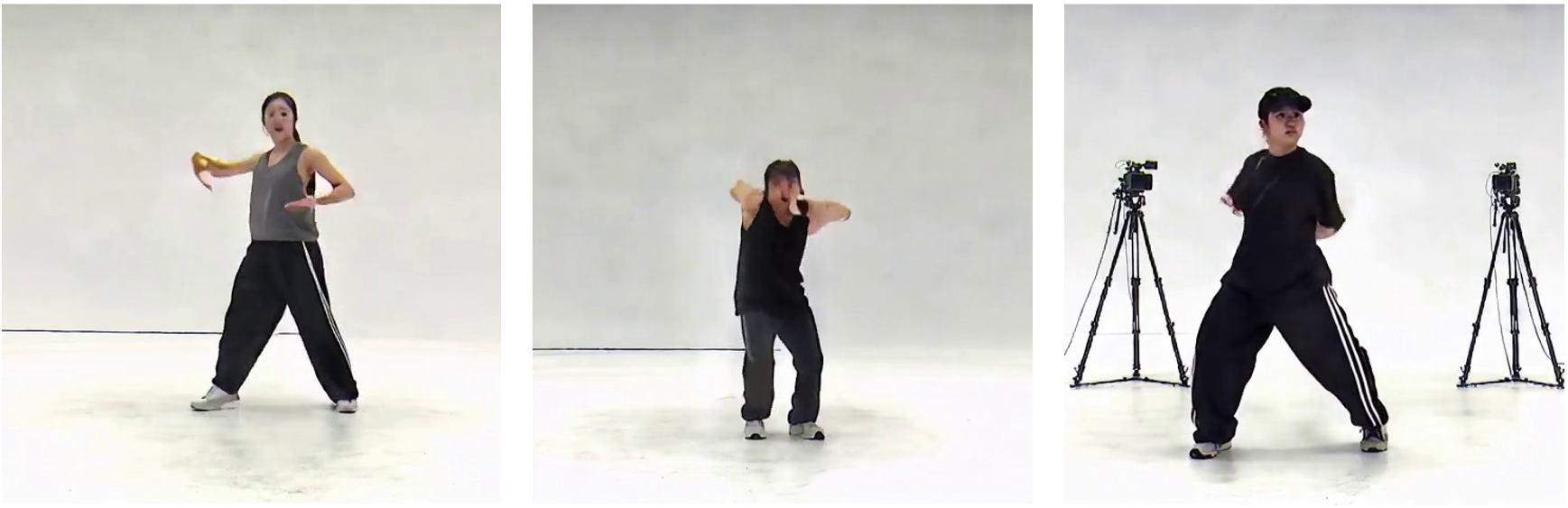}
\caption{Failure cases. Our model inherits some issues from the base model, such as failing to generate fine details (e.g., fingers and faces) and being fooled by the silhouette of the dancers.}
\label{fig:failure}
\end{figure}

Although our method adds listening capability to text-to-video models and improves dance generation, some properties such as style capture from the prompt and imaging quality are bounded by the capabilities of the underlying models. Additionally, our method inherits some problems from text-to-video models. Sometimes, fine details such as fingers and faces fail to be generated properly, especially when our model synthesizes dance videos with fast movements. Furthermore, our model is easily misled by the silhouette of the dancers, meaning that under the same silhouette, it may merge or swap the positions of body parts, which is also a problem in the base model. We include some examples of failure cases in Fig.~\ref{fig:failure}.

\section{Additional Qualitative Analysis}
\begin{figure}[t]
\centering
\includegraphics[width=\linewidth]{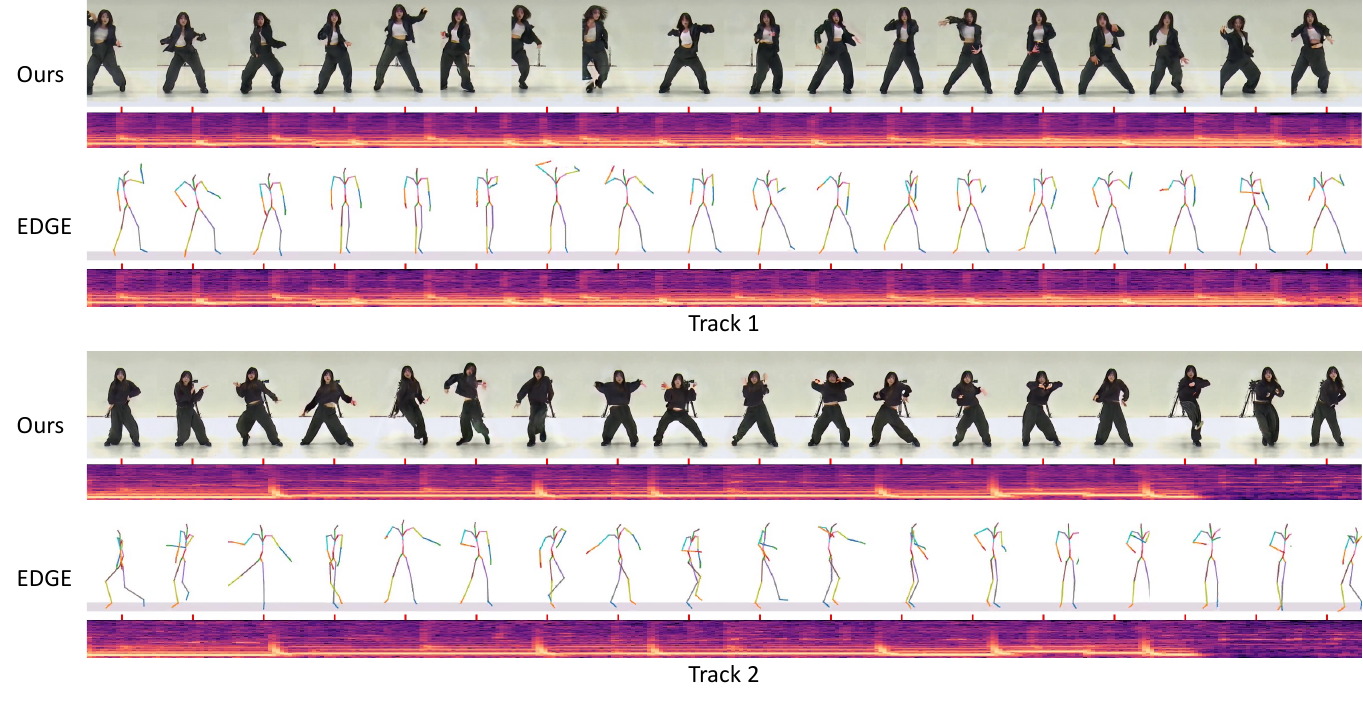}
\caption{Comparison with EDGE~\cite{tseng2023edge}. Note that the poses of our method and EDGE need not align.}
\label{fig:reb-edge}
\end{figure}

\begin{figure}[ht]
\center
\includegraphics[width=1.0\linewidth]{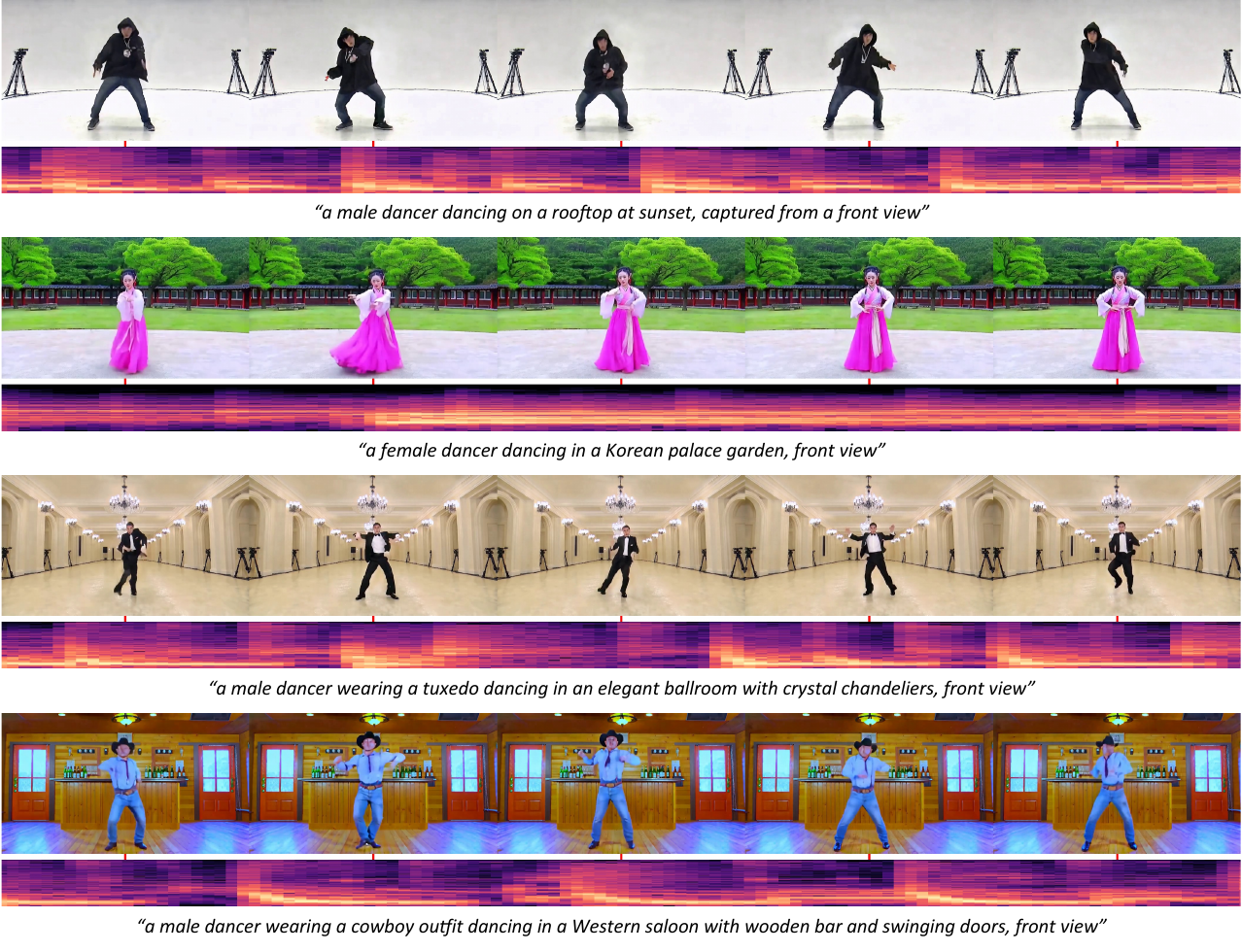}
\caption{More music-and-text-to-video generation results.}
\label{fig:ours-2}
\end{figure}

We show more music-and-text-to-video generation examples in Fig.~\ref{fig:ours-2}. Fig.~\ref{fig:mm-diffusion} presents a side-by-side comparison with MM-Diffusion~\cite{ruan2023mm}. Unlike MM-Diffusion, which generates shorter videos with limited style control, MusicInfuser produces longer sequences with both musical synchronization and prompt-based style control, while improving the overall consistency of the video and reducing artifacts. We show a comparison with Mochi~\cite{genmo2024mochi} in Fig.~\ref{fig:mochi-comparison}. Note that Mochi is not able to hear the music. Compared to Mochi, MusicInfuser produces more consistent human forms, fewer visual artifacts, and more fluid, realistic movements. Our method adds music responsiveness while maintaining or improving video consistency. We also compare with EDGE~\cite{tseng2023edge}, a state-of-the-art skeleton-based dance generation model. EDGE performs 3D skeleton generation requiring 3D pose reconstruction, whereas our method performs direct video synthesis. Fig.~\ref{fig:reb-edge} shows a visual comparison under the same music track using our method and EDGE. This demonstrates that our approach captures nuances that skeleton-based methods cannot represent, such as hair dynamics, clothing motion, a flexible backbone, and hand articulation.

We present qualitative results of our ablation study in Fig.~\ref{fig:abl-1} and Fig.~\ref{fig:abl-2}. Our full model successfully generates consistent body shapes that align with the music while preserving prior knowledge without introducing significant artifacts.

\section{Additional Quantitative Analysis}

Similar to the layer selection baselines and ablation studies in the main paper using Qwen3-Omni~\cite{xu2025qwen3}, we show evaluation using VideoLLaMA 2~\cite{cheng2024videollama} in Tables~\ref{tab:layer} and~\ref{tab:abl}. The full model achieves the highest score. Using a higher rank for LoRA contributes substantially to movement realism, while our Beta-Uniform scheduling improves body representation. The naive feature addition baseline, where instead of using the ZICA adapter we simply spatially expand the audio feature and add it to the corresponding frame, performs worse than our approach on most metrics, confirming the effectiveness of our ZICA strategy. In Table~\ref{tab:quality-comparison-refined}, we present comparisons between MM-Diffusion and our method, both trained on the identical AIST++ training dataset without in-the-wild data. This shows that our model trained on the AIST dataset alone already surpasses MM-Diffusion, while in-the-wild data further enhances generalization capability.

In Table~\ref{tab:prompt-alignment} in the main paper, we show the trade-off between style capture and creative interpretation of the prompt depending on the base prompt ratio, meaning how frequently we replaced the prompt with the basic prompt.

Additionally, we present commonly used intrinsic metrics from related work, BeatAlign and kinetic diversity~\cite{li2021ai,siyao2022bailando}, measured after extracting 2D pose sequences from generated videos. Table~\ref{tab:smpl-metrics} shows these metrics, demonstrating a comparable score to the AIST test set and superior scores compared to the baselines.

\section{Layer Adaptability}

\begin{figure}[t]
\center
\includegraphics[width=0.6\linewidth]{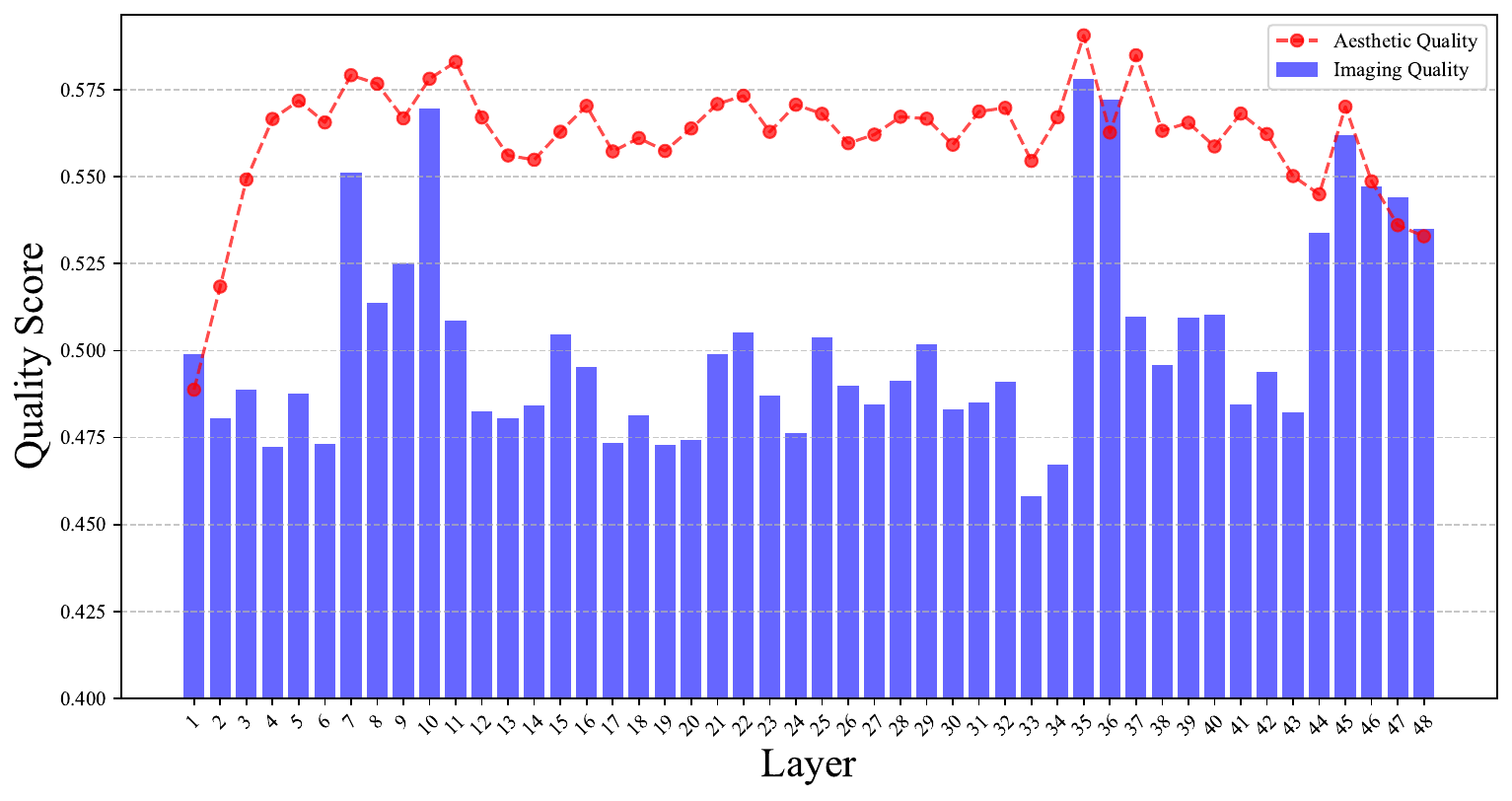}
\caption{Layer adaptability graph from \cite{hyung2024spatiotemporal}, showing imaging and aesthetic quality.}
\label{fig:layer}
\end{figure}

The imaging and aesthetic quality of the base model~\cite{genmo2024mochi} is presented in Fig.~\ref{fig:layer}. This is analyzed with STG~\cite{hyung2024spatiotemporal}, an inference-time technique, and the score is calculated with VBench~\cite{huang2024vbench}. Based on the imaging quality, which is highly related to the structure and noisiness of the video samples, we select the top 16 out of 48 layers in terms of imaging quality.

\begin{table}[t]
\centering
\scriptsize
\begin{tabular}{lcccccccc}
\toprule
\multirow{2}{*}{\textbf{Model}} & \textbf{Style} & \textbf{Beat} & \textbf{Body} & \textbf{Movement} & \textbf{Choreography} & \textbf{Imaging} & \textbf{Aesthetic} & \textbf{Overall} \\
& \textbf{Alignment} & \textbf{Alignment} & \textbf{Representation} & \textbf{Realism} & \textbf{Complexity} & \textbf{Quality} & \textbf{Quality} & \textbf{Consistency} \\
\midrule
MM-Diffusion~\cite{ruan2023mm} & 7.16 & 8.56 & 5.52 & 7.05 & 7.53 & 8.94 & 6.52 & 8.38 \\
Ours (Only AIST) & \textbf{7.83} & \textbf{9.10} & \textbf{6.89} & \textbf{8.58} & \textbf{7.96} & \textbf{9.55} & \textbf{8.02} & \textbf{9.75} \\
\bottomrule
\end{tabular}
\caption{Comparisons between MM-Diffusion~\cite{ruan2023mm} and our method, both trained on the AIST++ training dataset.}
\label{tab:quality-comparison-refined}
\end{table}

\begin{table}[t]
\centering
\scriptsize
\begin{tabular}{lcc}
\toprule
\textbf{Method} & \textbf{BeatAlign↑} & \textbf{Dist$_k$↑} \\
\midrule
{\color{gray}AIST Dataset (GT)} & {\color{gray}0.2448} & {\color{gray}9.027} \\
MM-Diffusion~\cite{ruan2023mm} & 0.1553 & 2.126 \\
Mochi~\cite{genmo2024mochi} & 0.1976 & 8.886 \\
MusicInfuser (Ours) & \textbf{0.2432} & \textbf{9.849} \\
\bottomrule
\end{tabular}
\caption{BeatAlign and kinetic diversity metrics based on 2D poses.}
\label{tab:smpl-metrics}
\end{table}

\section{Beta-Uniform Scheduling}

\begin{figure}[ht]
\center
\includegraphics[width=0.6\linewidth]{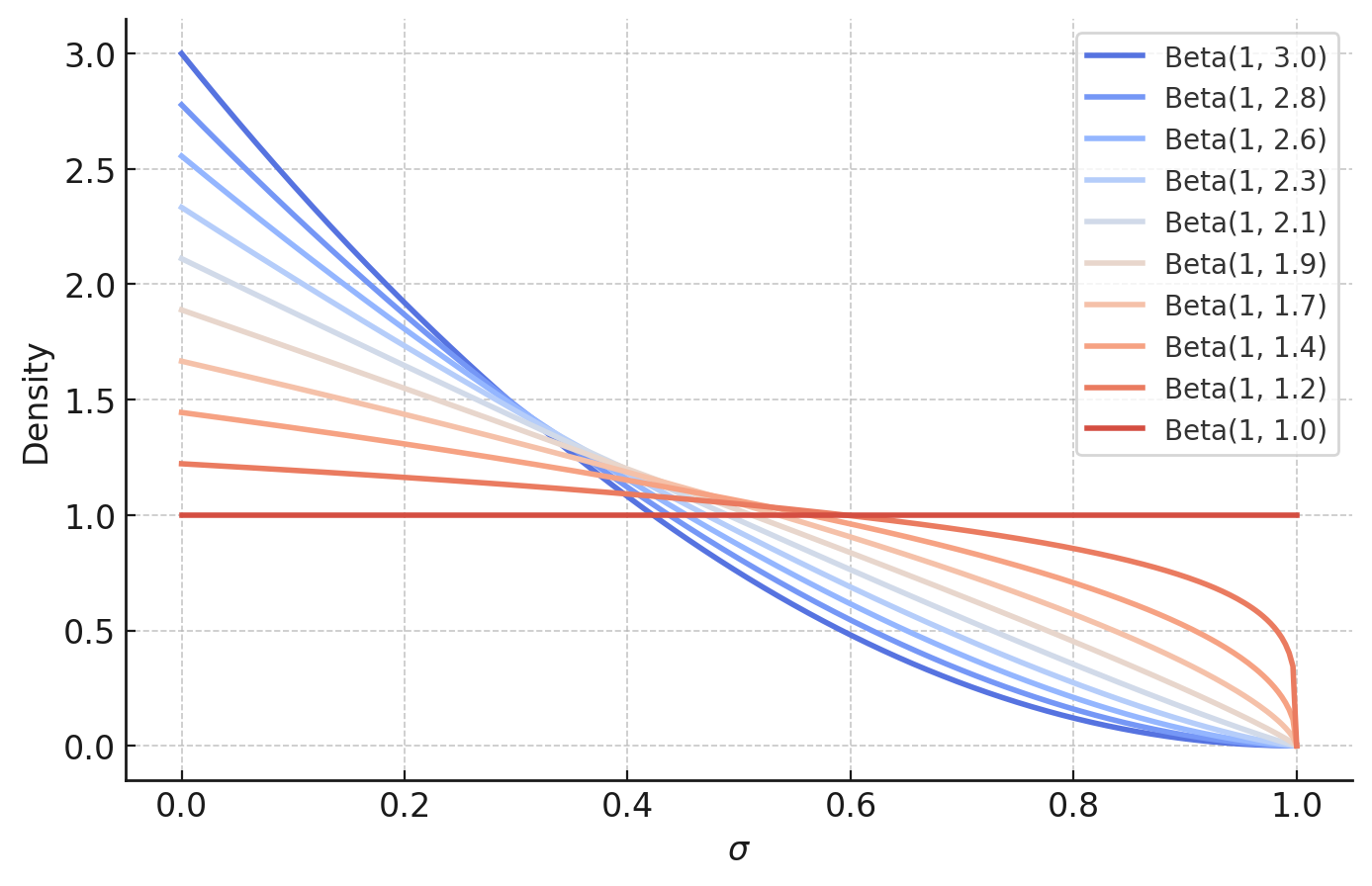}
\caption{Beta distributions.}
\label{fig:beta-uniform}
\end{figure}

The visualization of the Beta-Uniform scheduling strategy is shown in Fig.~\ref{fig:beta-uniform}.

\section{Test Music Tracks}

For evaluating our method, we use music tracks that are set aside from the training set~\cite{tsuchida2019aist}, following AIST++~\cite{li2021ai}. The full list of test music codes is provided in Table~\ref{tab:music-codes}.

\begin{table}[htbp]
\centering
\begin{tabular}{|c|l|}
\hline
\textbf{Test Music Code} & \textbf{Genre} \\
\hline
mLH4 & LA style Hip-hop \\
\hline
mKR2 & Krump \\
\hline
mBR0 & Break \\
\hline
mLO2 & Lock \\
\hline
mJB5 & Ballet Jazz \\
\hline
mWA0 & Waack \\
\hline
mJS3 & Street Jazz \\
\hline
mMH3 & Middle Hip-hop \\
\hline
mHO5 & House \\
\hline
mPO1 & Pop \\
\hline
\end{tabular}
\caption{List of test music codes with corresponding dance genres.}
\label{tab:music-codes}
\end{table}

\section{Prompts}

As mentioned in our main paper, we use a proper prompt format and base prompt for AIST~\cite{tsuchida2019aist}. The full list is shown in Table~\ref{tab:prompt-templates}. Note that since we use VideoChat2~\cite{li2023videochat} to label YouTube videos, we have only the base prompt for that dataset. We also provide a predefined set of prompts in Table~\ref{tab:dance-prompts} that is used to generate samples for the evaluation, ultimately resulting in $10\times 10 = 100$ videos per model configuration. The system prompts for VideoLLaMA 2~\cite{cheng2024videollama} and Qwen3-Omni~\cite{xu2025qwen3} used for evaluation are provided in Table~\ref{tab:eval-prompts}.

\begin{table}[htbp]
\centering
\begin{tabular}{|c|c|p{0.75\textwidth}|}
\hline
\textbf{Category} & \textbf{Dataset} & \textbf{Prompt Template} \\
\hline
Prompt Format & AIST & \{dancers\_text\} dancing \{genre\_name\} in a \{situation\_name\} setting in a studio with a white backdrop, captured from a \{camera\_view\} \\
\hline
Prompt Format & AIST & a \{camera\_view\} video of \{dancers\_text\} performing \{genre\_name\} choreography against a white background in a \{situation\_name\} scene \\
\hline
Prompt Format & AIST & \{dancers\_text\} executing \{genre\_name\} movements in a minimalist studio space in a \{situation\_name\} setting, shot from a \{camera\_view\} \\
\hline
Prompt Format & AIST & a \{genre\_name\} dance performance by \{dancers\_text\} in a pristine white studio, \{camera\_view\}, \{situation\_name\} \\
\hline
Base Prompt & AIST & a professional dancer dancing in a studio with a white backdrop \\
\hline
Base Prompt & YouTube & a dance video \\
\hline
\end{tabular}
\caption{Dance prompt templates categorized by type and dataset, including parameterized formats and simple base prompts.}
\label{tab:prompt-templates}
\end{table}

\begin{table}[htbp]
\centering
\begin{tabular}{|p{0.95\textwidth}|}
\hline
\textbf{Prompts} \\
\hline
a male dancer dancing on a rooftop at sunset, captured from a front view \\
\hline
a female dancer dancing in a subway station, captured from a front view \\
\hline
a male dancer dancing in an art gallery with some paintings, captured from a front view \\
\hline
a female dancer wearing a leather jacket dancing in a studio with a white backdrop, captured from a front view \\
\hline
a male dancer wearing a hoodie dancing in a studio with a white backdrop, captured from a front view \\
\hline
a female dancer wearing a denim vest dancing in a studio with a white backdrop, captured from a front view \\
\hline
a female dancer wearing a Hawaiian dress dancing on Waikiki Beach at sunset with Diamond Head in the background, captured from a front view \\
\hline
a male dancer wearing a suit dancing in the middle of a New York City, captured from a front view \\
\hline
a male dancer wearing a chef's uniform dancing in a busy restaurant kitchen with flames from the grill behind him, captured from a front view \\
\hline
a female dancer wearing a Renaissance gown dancing in a Venetian masquerade ball with ornate chandeliers overhead, captured from a front view \\
\hline
\end{tabular}
\caption{Collection of dance scene prompts with various subjects, attire, and settings.}
\label{tab:dance-prompts}
\end{table}

\begin{table}[htbp]
\centering
\begin{tabular}{|p{0.2\textwidth}|p{0.8\textwidth}|}
\hline
\textbf{Metric} & \textbf{Prompt} \\
\hline
\multicolumn{2}{|c|}{\textbf{Dance Quality}} \\
\hline
Style Alignment & Rate the style alignment of the dance to music where: 0 means poor style alignment of the dance to music, 5 means moderate style alignment of the dance to music, and 10 means perfect style alignment of the dance to music. Output only the number. \\
\hline
Beat Alignment & Rate the beat alignment of the dance to music where: 0 means poor beat alignment of the dance to music, 5 means moderate beat alignment of the dance to music, and 10 means perfect beat alignment of the dance to music. Output only the number. \\
\hline
Body Representation & Rate the body representation of the dancer where: 0 means unrealistic/distorted proportions of the dancer, 5 means minor anatomical issues of the dancer, and 10 means anatomically perfect representation of the dancer. Output only the number. \\
\hline
Movement Realism & Rate the movement realism of the dancer where: 0 means poor movement realism of the dancer, 5 means moderate movement realism of the dancer, and 10 means perfect movement realism of the dancer. Output only the number. \\
\hline
Choreography Complexity & Rate the complexity of the choreography where: 0 means extremely basic choreography, 5 means intermediate choreography, and 10 means extremely complex/advanced choreography. Output only the number. \\
\hline
\multicolumn{2}{|c|}{\textbf{Video Quality}} \\
\hline
Imaging Quality & Rate the imaging quality where: 0 means poor imaging quality, 5 means moderate imaging quality, and 10 means perfect imaging quality. Output only the number. \\
\hline
Aesthetic Quality & Rate the aesthetic quality where: 0 means poor aesthetic quality, 5 means moderate aesthetic quality, and 10 means perfect aesthetic quality. Output only the number. \\
\hline
Overall Consistency & Rate the overall consistency where: 0 means poor consistency, 5 means moderate consistency, and 10 means perfect consistency. Output only the number. \\
\hline
\multicolumn{2}{|c|}{\textbf{Prompt Alignment}} \\
\hline
Style Capture & How well does the dance video capture the specific style mentioned in the prompt: '\{prompt\}'? Rate 0-10 where: 0 means completely missed the style, 5 means some elements of the style are present, and 10 means perfectly captures the style. Output only the number. \\
\hline
Creative Interpretation & Based on the prompt '\{prompt\}', rate the creativity in interpreting the prompt 0-10 where: 0 means generic/standard interpretation, 5 means moderate creativity, and 10 means highly creative and unique interpretation. Output only the number. \\
\hline
Overall Prompt Satisfaction & Rate the overall prompt satisfaction 0-10 where: 0 means the video fails to satisfy the prompt '\{prompt\}', 5 means it partially satisfies the prompt, and 10 means it fully satisfies all aspects of the prompt. Output only the number. \\
\hline
\end{tabular}
\caption{System prompts for evaluation}
\label{tab:eval-prompts}
\end{table}

\section{Concurrent Work}

Several concurrent approaches have emerged alongside our research that address related challenges. Notable among these is VideoJAM~\cite{chefer2025videojam}, which enhances motion generation by jointly denoising both the motion maps and the video, an approach that is orthogonal to ours. Another related line of research is OmniHuman-1~\cite{lin2025omnihuman}, which integrates audio and pose inputs into diffusion models. The application of OmniHuman-1 remains primarily confined to scenarios that do not require much creative movement, relies on a private model, and necessitates full fine-tuning procedures, which distinguishes it from our approach.

